\def\GSSCarxiv{1}
\documentclass[journal]{IEEEtran}
\ifdefined\GSSCarxiv
  \def\GSSCpdfsubject{Preprint}
\else
  \def\GSSCpdfsubject{IEEE Transactions on Pattern Analysis and Machine Intelligence}
\fi
\usepackage{amsmath,amsfonts,amssymb,amsthm}
\DeclareMathOperator*{\argmax}{arg\,max}

\usepackage{array}
\usepackage{rotating}
\usepackage{multirow}
\newtheoremstyle{compactthm}{3pt}{3pt}{\itshape}{0pt}{\bfseries}{.}{ }{}
\theoremstyle{compactthm}

\theoremstyle{remark}

\usepackage[caption=false,font=normalsize,labelfont=sf,textfont=sf]{subfig}
\usepackage{textcomp}
\usepackage{xcolor}
\definecolor{darkamber}{RGB}{180,120,0}
\usepackage{stfloats}
\usepackage{url}
\usepackage{verbatim}
\usepackage{graphicx}
\usepackage{cite}
\usepackage{booktabs}
\usepackage{algorithm}

\usepackage{algpseudocode}
\usepackage{tikz}
\usetikzlibrary{arrows.meta,calc,patterns,positioning}
\usepackage[hidelinks]{hyperref}
\hypersetup{
  pdftitle={Generative Semantic Scene Completion},
  pdfauthor={Shi Chen and Weifeng Ge},
  pdfsubject={\GSSCpdfsubject},
  pdfkeywords={Semantic scene completion, discrete diffusion models, LiDAR point clouds,
               synthetic training data, bird's-eye view perception, autonomous driving,
               3D scene understanding},
}
\usepackage[capitalize]{cleveref}

\crefname{algorithm}{Algorithm}{Algorithms}
\Crefname{algorithm}{Algorithm}{Algorithms}
\ifdefined\GSSCarxiv
\else
  \usepackage{xr-hyper}
\fi
\usepackage{balance}
\usepackage{microtype}
\usepackage{placeins}

\let\origsection\section
\renewcommand{\section}{\FloatBarrier\origsection}

\newif\iftbdproof \tbdprooftrue

\newcommand{\na}{n/a}

\renewcommand{\arraystretch}{0.92}
\begin{document}

\title{Generative Semantic Scene Completion}

\author{Shi~Chen~and~Weifeng~Ge
\thanks{This work was supported by the National Natural Science Foundation of China under Grant~624B1006 and by the Shanghai Science and Technology Committee under Grant~24511103900. \textit{(Corresponding author: Weifeng Ge.)}}
\thanks{S. Chen and W. Ge are with the College of Computer Science and Artificial Intelligence, Fudan University, Shanghai, China. S. Chen is also with the Robotics Institute, Carnegie Mellon University, Pittsburgh, PA, USA.
E-mail: shichen22@m.fudan.edu.cn, shic@andrew.cmu.edu, wfge@fudan.edu.cn}
\thanks{Code, checkpoints, and the PS\textsuperscript{3} dataset are available at \protect\href{https://github.com/BillyChern/GSSC-S2D2}{\protect\nolinkurl{github.com/BillyChern/GSSC-S2D2}}.}
\thanks{Our SemanticKITTI leaderboard entry is at \protect\href{https://www.codabench.org/competitions/13814/}{\protect\nolinkurl{codabench.org/competitions/13814}}.}
}

\ifdefined\GSSCarxiv
\markboth{Preprint. Under review.}%
{Chen and Ge: Generative Semantic Scene Completion}
\else
\markboth{IEEE Transactions on Pattern Analysis and Machine Intelligence,~Vol.~XX, No.~X, Month~Year}%
{Chen and Ge: Generative Semantic Scene Completion}
\fi

\maketitle

\begin{abstract}
Outdoor LiDAR semantic scene completion (SSC) recovers a dense semantic voxel grid from a scan observing 1\% of the target volume, under class imbalance beyond 7{,}000$\times$. We recast SSC as generative semantic scene completion (GSSC): a single discrete-diffusion formulation in three roles. First, paired sparse–dense scene synthesis (PS\textsuperscript{3}) generates matched sparse LiDAR observations with their dense semantic completions, addressing the long tail at its source and yielding the PS\textsuperscript{3}-SemanticKITTI corpus we train on alongside SemanticKITTI. Second, semantic-guided generative scene completion (SGSC) generates the scene from noise with multinomial discrete diffusion, conditioned on the sparse scan through a bird's-eye-view semantic map and a sparse 3D feature stream. Third, the same framework instead refines an existing completion in one flow-matching step: structured source discrete diffusion (S\textsuperscript{2}D\textsuperscript{2}). S\textsuperscript{2}D\textsuperscript{2} improves the mIoU of SGSC's own output and every external SSC base tested, without base retraining or test-time adaptation. On the strongest base, one step without test-time augmentation reaches $\mathbf{38.8\%}$ mIoU on the SemanticKITTI hidden test. To our knowledge that is the best \emph{causal, single-sweep, single-sample} result on that leaderboard, $+2.1$\,pp over the previous best published score under the same restriction. Four correction steps with eight-view test-time augmentation reach $39.2\%$, outside that restriction.\end{abstract}

\begin{IEEEkeywords}
Semantic scene completion, discrete diffusion models, LiDAR point clouds, synthetic training data, bird's-eye view perception, autonomous driving, 3D scene understanding
\end{IEEEkeywords}

\section{Introduction}
\label{sec:intro}

\begin{figure*}[t]
    \centering
    \begin{minipage}[t]{0.41\textwidth}
        \vspace{0pt}
        \centering
        \includegraphics[width=\textwidth]{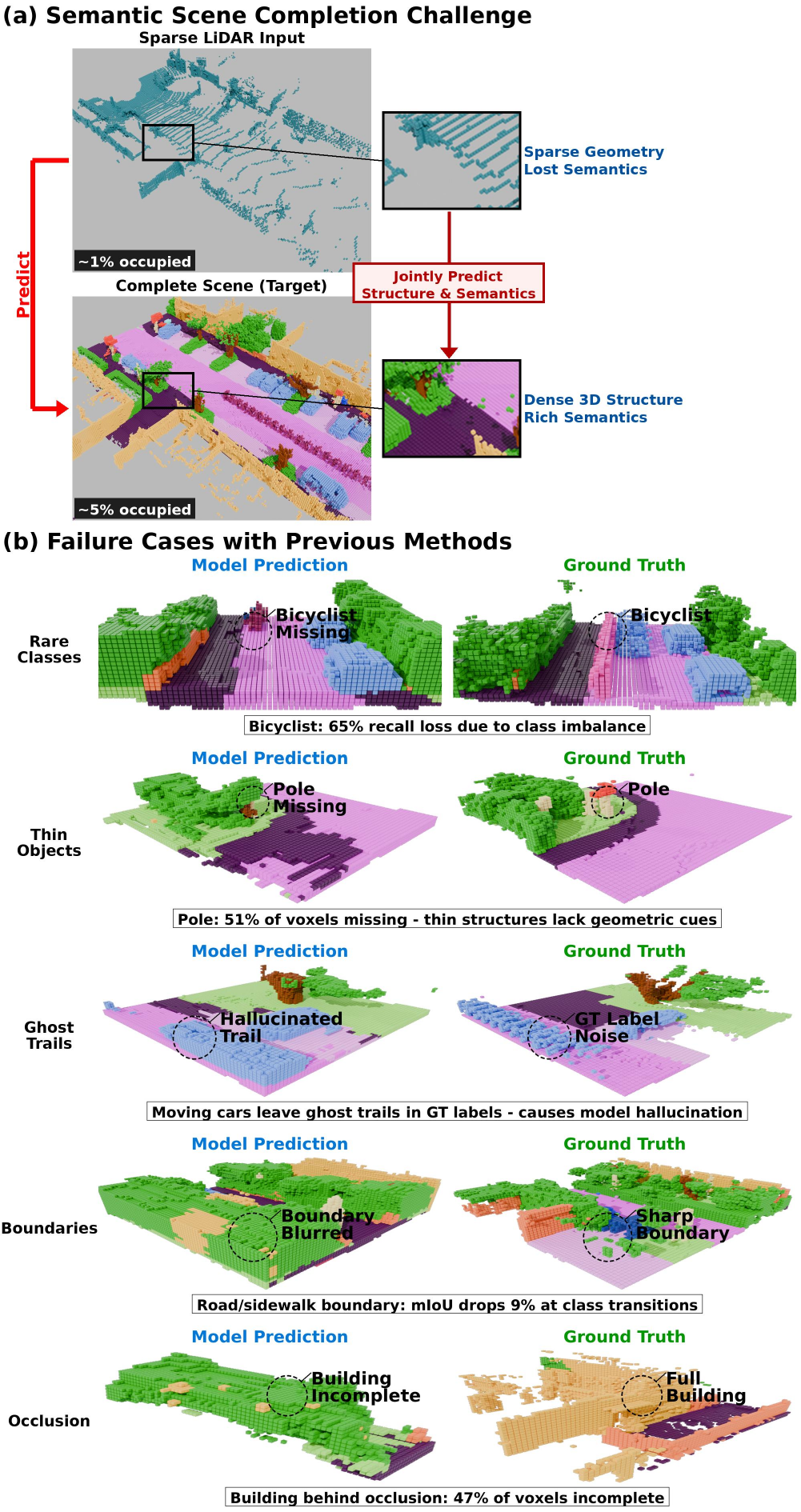}
    \end{minipage}
    \hspace{0.012\textwidth}
    \begin{minipage}[t]{0.3070\textwidth}
        \vspace{0pt}
        \centering
        \includegraphics[width=\textwidth]{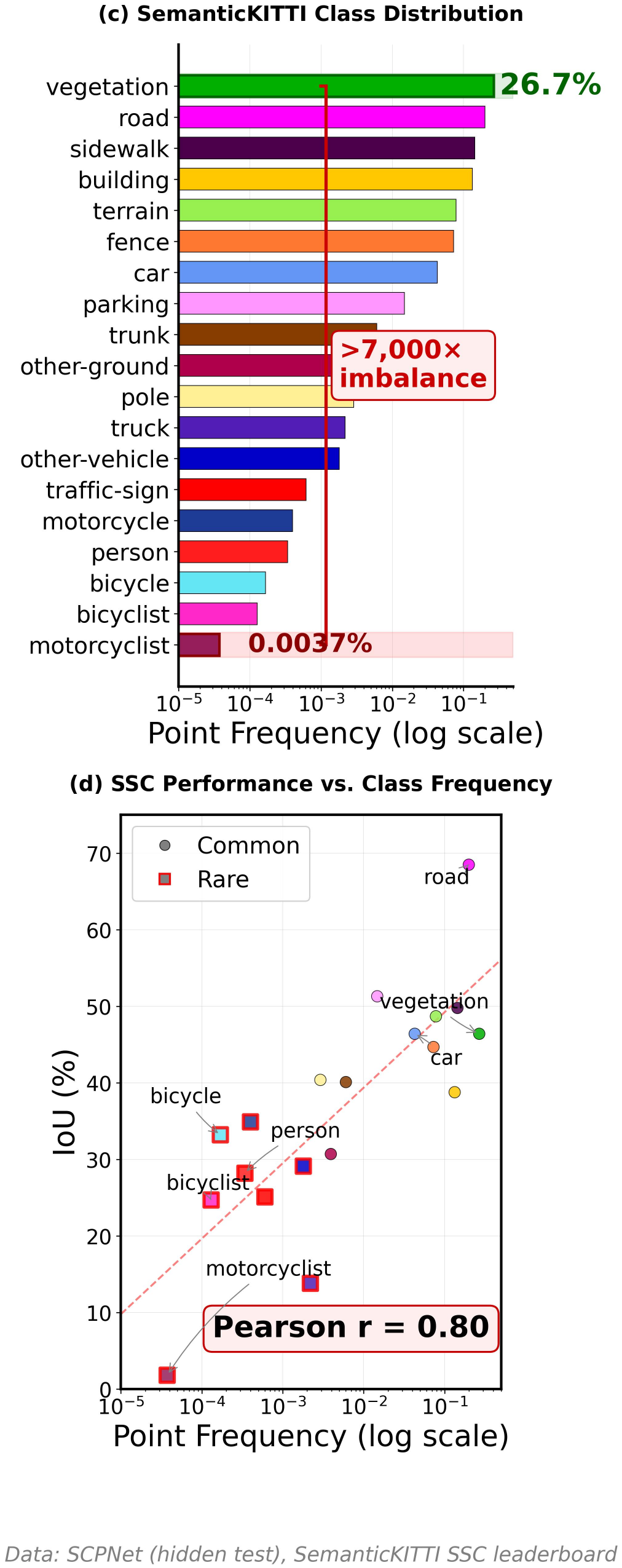}
    \end{minipage}
    \caption{\textbf{The challenge of semantic scene completion.} (a)~SSC infers dense structure and semantics from a sparse sweep. (b)~Five failure modes of discriminative methods; the percentages are measured on the validation predictions of SCPNet~\cite{xia2023scpnet}. (c)~SemanticKITTI is severely long-tailed, \emph{vegetation} $26.7\%$ of labelled points against \emph{motorcyclist} $0.0037\%$. (d)~Our class frequency (log axis) and SCPNet's published hidden-test IoU correlate at Pearson $r{=}0.80$.}
    \label{fig:teaser}
\end{figure*}

\IEEEPARstart{S}{emantic} scene completion (SSC) is a key element of 3D perception for autonomous driving. A vehicle cannot plan against the surfaces its sensor happens to see: it must reason about the space behind a parked car and the road that continues past the last returned point~\cite{hu2023planning,jiang2023vad}. Supervised networks learn to supply that volume because one benchmark labels it. SemanticKITTI~\cite{behley2019semantickitti} aggregates a whole sequence of scans into each target grid, so a voxel occluded in the current frame is still labelled if a later scan reached it. That construction also fixes the class histogram to whatever the recording vehicle drove past. Vegetation outnumbers motorcyclist by more than three orders of magnitude, and per-class accuracy follows. Our class frequency and the previous method's published hidden-test per-class IoU correlate strongly~\cite{xia2023scpnet}, so the classes a planner most needs are the ones the training set supplies least. The sensor compounds this. One sweep returns about one voxel in a hundred, and everything behind, inside and beyond what it struck has to be inferred, which is where rare classes, thin structures, ghost trails, boundary errors and occlusion all surface (\cref{fig:teaser}).

Two remedies address the imbalance from inside that corpus. The first reweights the objective (class weights, or focal loss~\cite{lin2017focal}), which changes how much each existing example counts, not how many exist. The second extracts more supervision from the same sequences (\cref{sec:related}). Neither adds a scene, so both work against a histogram that does not move.

Recognition and completion are not two stages for a person: we label what is visible while hallucinating what is not, the perceptual filling-in that Gestalt psychology named \emph{amodal completion}~\cite{michotte1964amodal,kanizsa1979organization}. Feynman's ``What I cannot create, I do not understand'' states what that asks of a model. A discriminative network predicts labels for what it sees; a generative one has to learn the structure of what could exist.

Generating 3D data is no longer the obstacle (\cref{sec:related}), but the two families that could supply one each leave a gap. An unconditional generator samples a complete labelled grid from noise, with no sparse observation to be completed from and the corpus histogram intact. A conditional one completes from noise, rebuilding the ground plane and the road each time even where a deterministic prediction already places them.

We propose to close both gaps with a single discrete-diffusion formulation in three roles, and call the framework generative semantic scene completion (GSSC). Paired sparse--dense scene synthesis (PS\textsuperscript{3}) gives the generated scene an observation. Pyramid discrete diffusion synthesises the complete grid, a rare-class object bank pastes in the classes the generator seldom places, and a Velodyne HDL-64E ray-casting model resamples the result into the sparse scan that would have produced it. The amplification a class receives therefore follows the bank's supply rather than the class's rarity. Semantic-guided generative scene completion (SGSC) covers the case where no base network exists to start from, diffusing from pure categorical noise conditioned on bird's-eye-view semantics and frozen LiDAR-segmentation features read off the sweep. Structured source discrete diffusion (S\textsuperscript{2}D\textsuperscript{2}) replaces that noise source with a frozen network's own prediction and learns a deterministic transport from it to the ground truth, so nothing is rediscovered and one step suffices at deployment.

Alongside the framework we report the analyses an adopter would need. On the SemanticKITTI hidden test one sampling step with no test-time augmentation reaches $38.8\%$ mIoU, $+2.1$\,pp over SCPNet's published $36.7\%$, indexed under a predicate that fixes what one prediction may use: one sweep, no future moments, no ensembling. That predicate excludes multi-sweep entries (SCPNet at four sweeps publishes $47.5\%$), test-time adaptation (TALoS reaches $37.9\%$ by tuning the forward pass's weights on other moments, future ones included), and a four-step, eight-view ensemble of our own that would otherwise read $39.2\%$. We refine three frozen bases spanning the field without retraining any, train on the synthetic pool, and test the frozen checkpoint zero-shot on two further datasets (\cref{sec:experiments}).

In sum, our main contributions are the following:

\begin{itemize}
    \item We recast semantic scene completion as generative modelling, driven by a single discrete-diffusion formulation that synthesises training data (PS\textsuperscript{3}), completes scenes (SGSC), and refines existing completions (S\textsuperscript{2}D\textsuperscript{2}).

    \item \looseness=-1 Refining a frozen network's prediction in a single step, with no retraining, posts, to our knowledge, the best causal, single-sweep, single-sample result on the SemanticKITTI hidden test.

    \item We give the exactness identity and the non-amplifying error bound behind one-step deployment---viable rather than provably sufficient---and state where each stops short of the shipped schedule.

\item Our paired synthesis yields PS\textsuperscript{3}-SemanticKITTI, a synthetic corpus of sparse--dense pairs with amplified rare classes, which improves completion on the voxel-grid-native base we test when trained alongside the real data (\cref{sec:synth_quality}).

    \item We identify the gaps to be addressed next: thin structures barely move (\cref{sec:perclass}), and no experiment here isolates occlusion.
\end{itemize}

\noindent Every name and metric we coin is also defined together on page~1 of the supplementary material.

\section{Related Work}
\label{sec:related}

SSC predicts occupancy and semantics jointly across a volume the sensor barely observes. Song~\textit{et al.}~\cite{song2017semantic} posed it indoors, from a single depth image. We group the outdoor work by \emph{what} each method changes, and treat the generative approaches separately below.

\noindent\textbf{Discrete diffusion.} \looseness=-1 The variant this paper builds on corrupts categorical variables rather than continuous ones. D3PM~\cite{austin2021structured} and the multinomial formulation of Hoogeboom~\textit{et al.}~\cite{hoogeboom2021argmax} replace Gaussian noise with a transition matrix over a label alphabet, so a voxel's class is resampled rather than perturbed. A categorical process therefore fits a semantic scene where a continuous one must quantise at the end. An instance is fixed by its \emph{source} and its \emph{kernel} (\cref{sec:prelim}); conditioning the reverse process on an observation turns generation into completion.

\noindent\textbf{Changing the network.} \looseness=-1 One line treats completion as voxel labelling, differing mainly in representation: a trunk without 3D convolution~\cite{roldao2020lightweight}, or a continuous field of local deep implicit functions in place of voxels~\cite{rist2021semantic}. Further variants fuse a bird's-eye-view~completion with a sparse-voxel one~\cite{cheng2021s3cnet}, merge a 3D segmentation branch into a bird's-eye-view~completion branch~\cite{yang2021ssasc}, couple completion to a segmentation backbone through point--voxel interaction~\cite{yan2021sparse}, or separate semantics from geometry~\cite{mei2023sscrs}. No representation changes what the sensor gives: about one voxel in a hundred, and unlabelled, so geometry must be inferred almost everywhere and semantics everywhere.

\noindent\textbf{Changing the supervision.} A second line changes what the network learns from: SCPNet~\cite{xia2023scpnet} distils a multi-frame teacher, TALoS~\cite{jang2024talos} adapts at test time under line-of-sight constraints, and PaSCo~\cite{cao2024pasco} adds instance supervision. Each gain is written into the base network's own weights.


\noindent\textbf{Changing the data.} A fourth line changes what the model trains on: CarlaSC~\cite{wilson2022motionsc} collects paired scans in a simulator. Pyramid discrete diffusion~\cite{liu2024pyramid} generates outdoor scenes outright. Neither meets a long-tailed benchmark's needs: CarlaSC's scenes are CARLA's~\cite{dosovitskiy2017carla}, not SemanticKITTI's, and an unconditional generator reproduces the distribution it was trained on, so it too seldom places the rare classes, and it supplies no matching sparse scan. Two nearer lines exist: discrete latent representations of LiDAR serve both completion and the generation of new scans~\cite{xiong2023compact}, and pasting instances between scans is long-standing practice in LiDAR training~\cite{yan2018second,xiao2022polarmix}. Our object bank belongs to that lineage, carried to scene scale and into the voxel label space.

In contrast to previous works, we do not seek a better representation, richer supervision written into a base network's own weights, or an input built for dense image features. We learn a distribution over complete labelled scenes and read all three jobs off it, under two sources and three conditioning designs. Spanning several of those jobs from one generative model is not unique to us: Octree Diffusion~\cite{zhang2026octree} generates, completes and extends semantic scenes from one octree latent, conditioned on partial scans at inference without retraining. Ours is the third job, correcting a frozen base inside the label space it is scored in. The fourth axis we occupy from the other end: rather than collecting scenes we generate them, together with the observation that would have produced them, which is also LidarDM's~\cite{zyrianov2025lidardm} architecture. PS\textsuperscript{3} differs in purpose, supplying paired supervision with the rare classes amplified.

\looseness=-1 \noindent\textbf{Diffusion for 3D scenes.} Object-level 3D diffusion is mature: Point-Voxel Diffusion~\cite{zhou20213d}, LION~\cite{vahdat2022lion} and the point-cloud diffusion of Luo and Hu~\cite{luo2021diffusion} generate object shapes. At scene scale, DiffuScene~\cite{tang2023diffuscene} synthesises indoor layouts, and LiDM~\cite{ran2024towards} diffuses an outdoor scene in a latent space. In outdoor semantic scene completion, DiffSSC~\cite{cao2024diffssc} and OccGen~\cite{wang2024occgen} run continuous Gaussian diffusion to \emph{produce} a scene, leaving the label space; Lee~\textit{et al.}~\cite{lee2023diffusion} stay in it and complete a scene from a sparse point cloud.

The from-noise setting is therefore not ours to claim: that work already reaches it inside the label space. What SGSC changes is the conditioning. Where they read the sparse cloud, we read the bird's-eye-view semantics projected from it together with a sparse 3D feature stream. We report what each stream is worth under an oracle BEV rather than only in the regime we deploy.

Two gaps run through the outdoor-completion methods above: none supplies the matching sparse scan a completion pair needs, and none refines a scene at the resolution and label space it will be scored in. Refining an existing prediction is not itself new: SemCity~\cite{lee2024semcity} enhances completion outputs through a \emph{continuous} triplane diffusion prior where we diffuse the discrete field itself, LiDiff~\cite{nunes2024scaling} completes a single scan by diffusing points. ESSC-RM~\cite{zhang2026esscrm} appends a refinement module to camera-based backbones, and OccFiner~\cite{shi2024occfiner} refines camera occupancy offboard from several frames, outside the predicate we index on. PS\textsuperscript{3} closes the first, driving the pyramid generator~\cite{liu2024pyramid}, pasting in the rare classes a generator seldom places, and resampling the result into the scan that would have observed it; S\textsuperscript{2}D\textsuperscript{2} closes the second by taking an existing prediction as its source.

\noindent\textbf{Diffusion from a structured source.} Diffusion is not solely generative. One use corrects a model's output, as RefineCatDiff~\cite{liu2025refinecatdiff} refines an initial-stage segmentation in medical imaging. Replacing the noise prior with a data-dependent endpoint is established: ResShift~\cite{yue2023resshift} diffuses along an image residual; bridge models such as I\textsuperscript{2}SB~\cite{liu2023i2sb}, BBDM~\cite{li2023bbdm} and DDBM~\cite{zhou2024ddbm} transport between two given endpoints. PDE-Refiner~\cite{lippe2024pderefiner} renoises a solver's own earlier prediction, and Cold Diffusion~\cite{bansal2022cold} shows generation survives degradations that are not noise at all, the deterministic regime that S\textsuperscript{2}D\textsuperscript{2} inherits. InDI~\cite{delbracio2023inversion} iterates directly from the degraded input rather than from noise, on continuous images.

In the discrete setting, SegRefiner~\cite{wang2023segrefiner} runs a diffusion whose terminal state is a base model's coarse mask, and transports away from that mask stochastically. Discrete flow matching~\cite{gat2024discrete,campbell2024generalized} brings flow matching to categorical data, also with a stochastic kernel, and starts from a mask or uniform source distribution.

In this work we set the diffusion source to a frozen network's prediction and read the corruption as a deterministic interpolant between the two endpoints. Training still draws categorically around that path, but the sampler runs on the path itself, so the reverse transports the source toward the ground truth instead of rebuilding the scene around it. Determinism is what makes the error bound telescope rather than compound. At deployment it collapses onto one residual plus a small schedule offset, which makes a single Euler step viable rather than provably sufficient. The operator trains across the full schedule and deploys at one, so no distillation stage stands between training and inference, and it leaves every base weight untouched, so the gain cannot be a retrained base in disguise. Whether the multi-step path buys the gain is a separate question: our control holds the loss fixed but still cannot separate the schedule from the variational term it carries, so we scope that claim to the path as a whole. What the operator learns is the correction, not the scene.

\section{Methodology}
\label{sec:method}

Semantic scene completion aims to predict a complete 3D semantic voxel grid $\mathbf{Y} \in \{0, 1, \ldots, K-1\}^{L \times W \times H}$ from sparse LiDAR observations $\mathbf{X} \in \{0, 1\}^{L \times W \times H}$, where $K{=}20$ counts $19$ semantic classes plus one empty/unlabelled class and $(L, W, H) = (256, 256, 32)$ for SemanticKITTI~\cite{behley2019semantickitti}. Whereas prior methods regress a single deterministic completion, we learn the distribution of plausible complete scenes and generate completions by structured sampling. The GSSC framework instantiates one discrete-diffusion formulation at three stages spanning two sources---uniform noise and a frozen prediction---and three conditioning designs; SGSC and S\textsuperscript{2}D\textsuperscript{2} share the denoiser architecture but not its weights, and PS\textsuperscript{3} runs its own pyramid networks. The three follow in turn. PS\textsuperscript{3} (\cref{sec:data_aug}) builds the paired data the other two train on; SGSC (\cref{sec:sgsc}) completes a scene from noise under LiDAR-derived conditioning; and S\textsuperscript{2}D\textsuperscript{2} (\cref{sec:s2d2}) corrects a frozen base's prediction in one deterministic step, whatever model produced it.

\subsection{Preliminaries}
\label{sec:prelim}

Everything in this subsection is inherited; \cref{sec:sgsc,sec:s2d2} state what we change. We embed a
semantic grid onto the product simplex $\mathcal{M} = \prod_v \Delta^{K-1}$, sending a label $k$ to the
basis vector $\mathbf{e}_k$, where $\Delta^{K-1} := \{\mathbf{p}\in\mathbb{R}^K : p_k\ge 0,\ \sum_k p_k
= 1\}$. Multinomial diffusion~\cite{hoogeboom2021argmax} mixes each per-voxel label toward uniform
categorical noise, with $\mathbf{u} = \tfrac{1}{K}\mathbf{1}$ the uniform simplex point, $\{\beta_t\}_{t
=1}^{T}$ a mixing schedule, and cumulative retention $\bar\alpha_t := \prod_{r=1}^{t}(1-\beta_r)$. A
single step keeps the label with probability $1-\beta_t$ and otherwise resamples it uniformly,
$q(\mathbf{x}_t \mid \mathbf{x}_{t-1}) = \mathrm{Cat}(\mathbf{x}_t \mid (1-\beta_t)\mathbf{x}_{t-1} +
\beta_t\mathbf{u})$, with closed-form marginal $q(\mathbf{x}_t \mid \mathbf{x}_0) =
\mathrm{Cat}(\mathbf{x}_t \mid \bar\alpha_t\mathbf{x}_0 + (1-\bar\alpha_t)\mathbf{u})$. Since
$\bar\alpha_0\!=\!1$ and $\bar\alpha_T\!\approx\!0$, the terminal state $\mathbf{x}_T \approx
\mathbf{u}$ is the inference prior.

Bayes' rule along the chain gives the reverse posterior conditioned on the clean scene,
\begin{equation}
    q(\mathbf{x}_{t-1} \mid \mathbf{x}_t, \mathbf{x}_0) \;=\; \frac{q(\mathbf{x}_t \mid \mathbf{x}_{t-1})\, q(\mathbf{x}_{t-1} \mid \mathbf{x}_0)}{q(\mathbf{x}_t \mid \mathbf{x}_0)},
    \label{eq:sgsc_posterior}
\end{equation}
again categorical and, for $t>1$, strictly positive and available in closed form; at $t{=}1$ it
degenerates to the Dirac at $\mathbf{x}_0$. Since $\mathbf{x}_0$ is unobserved at inference, the
standard parameterisation~\cite{hoogeboom2021argmax} substitutes the denoiser's simplex-valued estimate
$\hat{\mathbf{x}}_0 := \hat{p}_\theta(\cdot\mid\mathbf{x}_t, t, \mathbf{c}) \in \Delta^{K-1}$,
\begin{equation}
    p_\theta(\mathbf{x}_{t-1} \mid \mathbf{x}_t, \mathbf{c}) \;:=\; q(\mathbf{x}_{t-1} \mid \mathbf{x}_t, \hat{\mathbf{x}}_0).
    \label{eq:sgsc_reverse}
\end{equation}
The timestep conditioning we use throughout, adaptive group normalisation~\cite{dhariwal2021diffusion},
is likewise standard. Flow Matching~\cite{lipman2023flow}, which \cref{sec:s2d2} builds on, is likewise inherited: it fits a velocity field to the constant displacement of the straight interpolant $\mathbf{z}_s=(1{-}s)\mathbf{z}_0 + s\mathbf{z}_1$ between a source and a target, minimising $\mathbb{E}\lVert\mathbf{v}_\theta(\mathbf{z}_s,s)-(\mathbf{z}_1-\mathbf{z}_0)\rVert^2$. The minimiser is the marginal velocity carrying $p_0$ to $p_1$.

\subsection{Paired Sparse--Dense Scene Synthesis (\texorpdfstring{PS\textsuperscript{3}}{PS3})}
\label{sec:data_aug}
\label{sec:ps3_dataset}
\label{sec:object_bank}

\begin{figure*}[!tbp]
    \centering
    \includegraphics[width=0.80\textwidth]{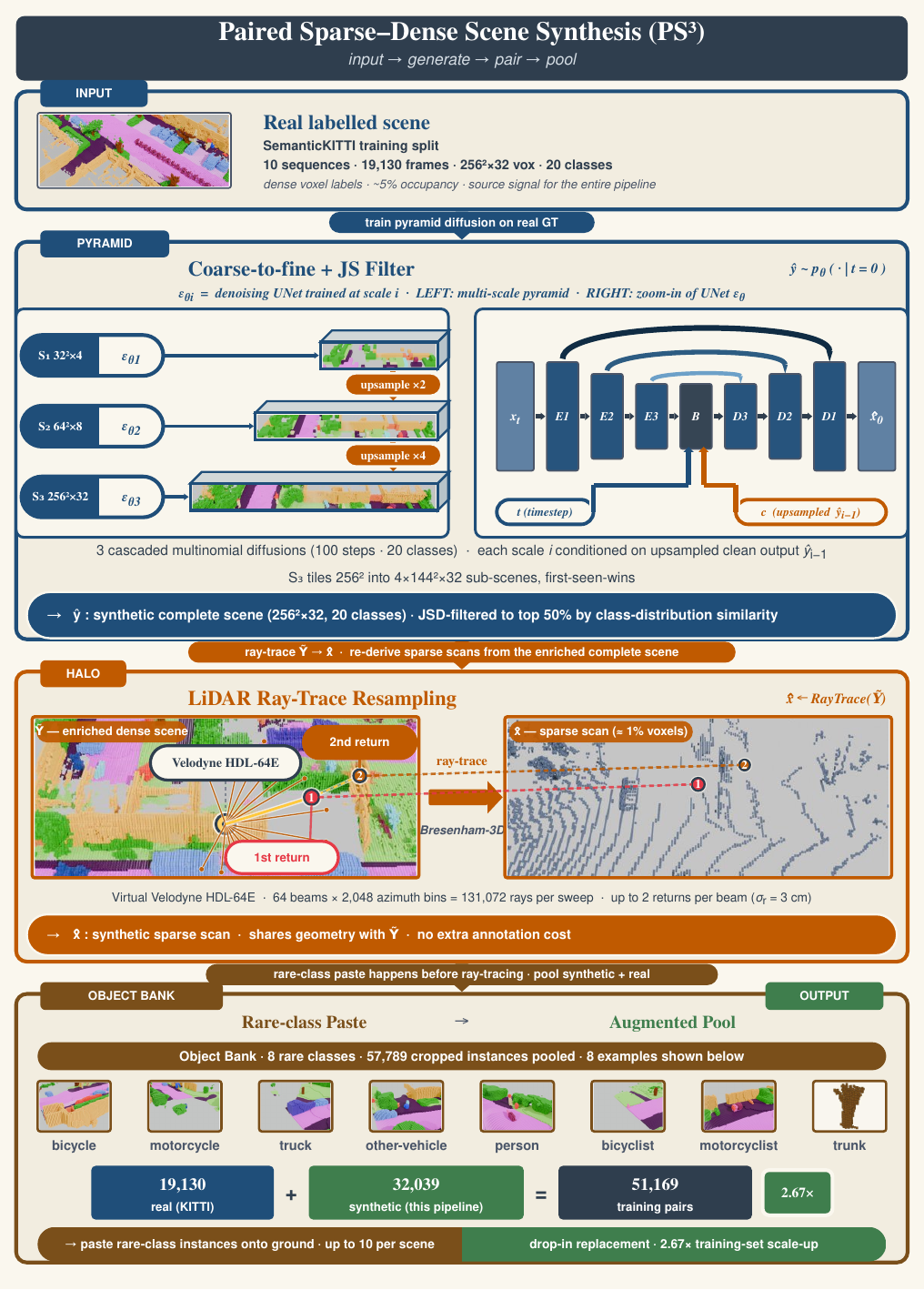}
    \caption{\textbf{PS\textsuperscript{3}: the offline data-augmentation pipeline.} Three cascaded coarse-to-fine multinomial diffusions, screened by a Jensen--Shannon divergence filter; a rare-class object bank and the Hardware-Aware LiDAR Observation (\textsc{Halo}) HDL-64E ray-tracer then turn each synthetic scene into a (sparse, dense) pair. Calibration, thresholds and the pasting pseudocode are in Appendix~A of the supplementary material.}
    \label{fig:ps3_pipeline}
\end{figure*}

\begin{figure*}[!tb]
    \centering
    \resizebox{0.80\textwidth}{!}{%
    \begin{tikzpicture}[
        font=\footnotesize,
        panel/.style={inner sep=0pt, outer sep=0pt},
        flow/.style={-{Stealth[length=2.6mm,width=2.4mm]}, line width=1.1pt, draw=black!62},
        badge/.style={font=\footnotesize\sffamily, fill=black!7, rounded corners=1.5pt, inner sep=2.2pt, text=black!80},
        sub/.style={font=\footnotesize, align=center, text=black!85},
    ]
        \def\PW{5.2cm}
        \node[panel] (s1) at (0,5.15)    {\includegraphics[width=\PW,trim=0 30 0 330,clip]{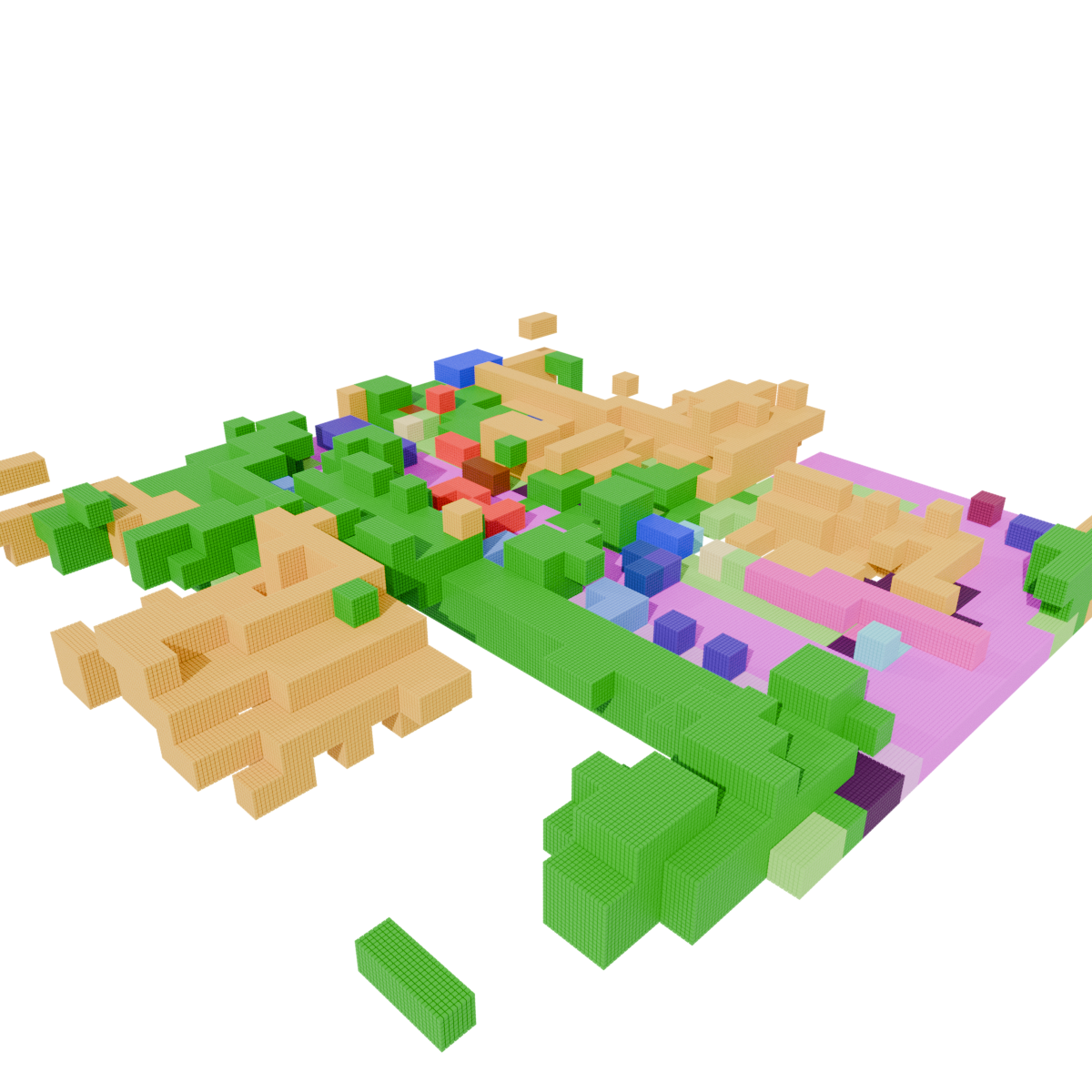}};
        \node[panel] (s2) at (5.95,5.15) {\includegraphics[width=\PW,trim=0 30 0 330,clip]{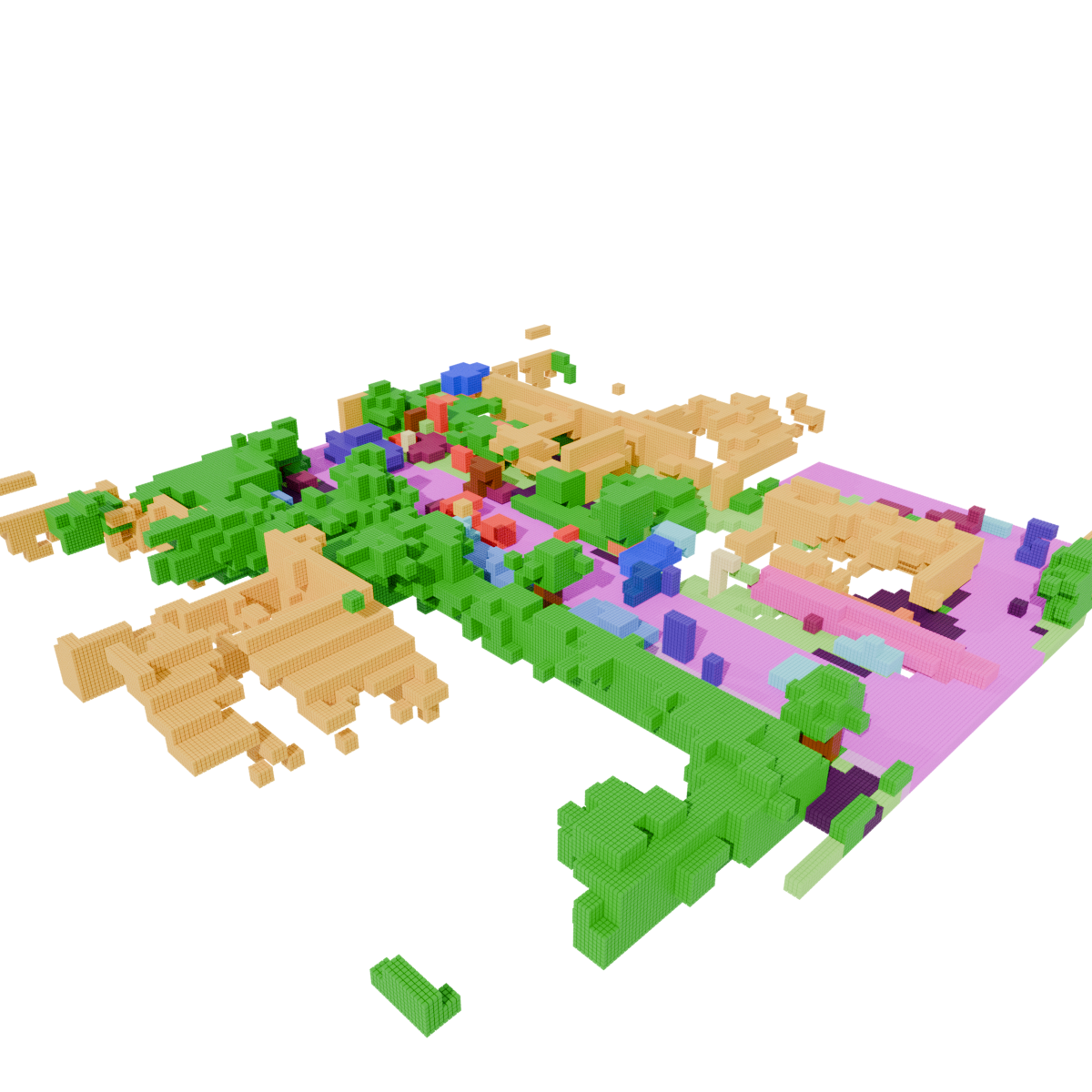}};
        \node[panel] (s3) at (11.9,5.15) {\includegraphics[width=\PW,trim=0 30 0 330,clip]{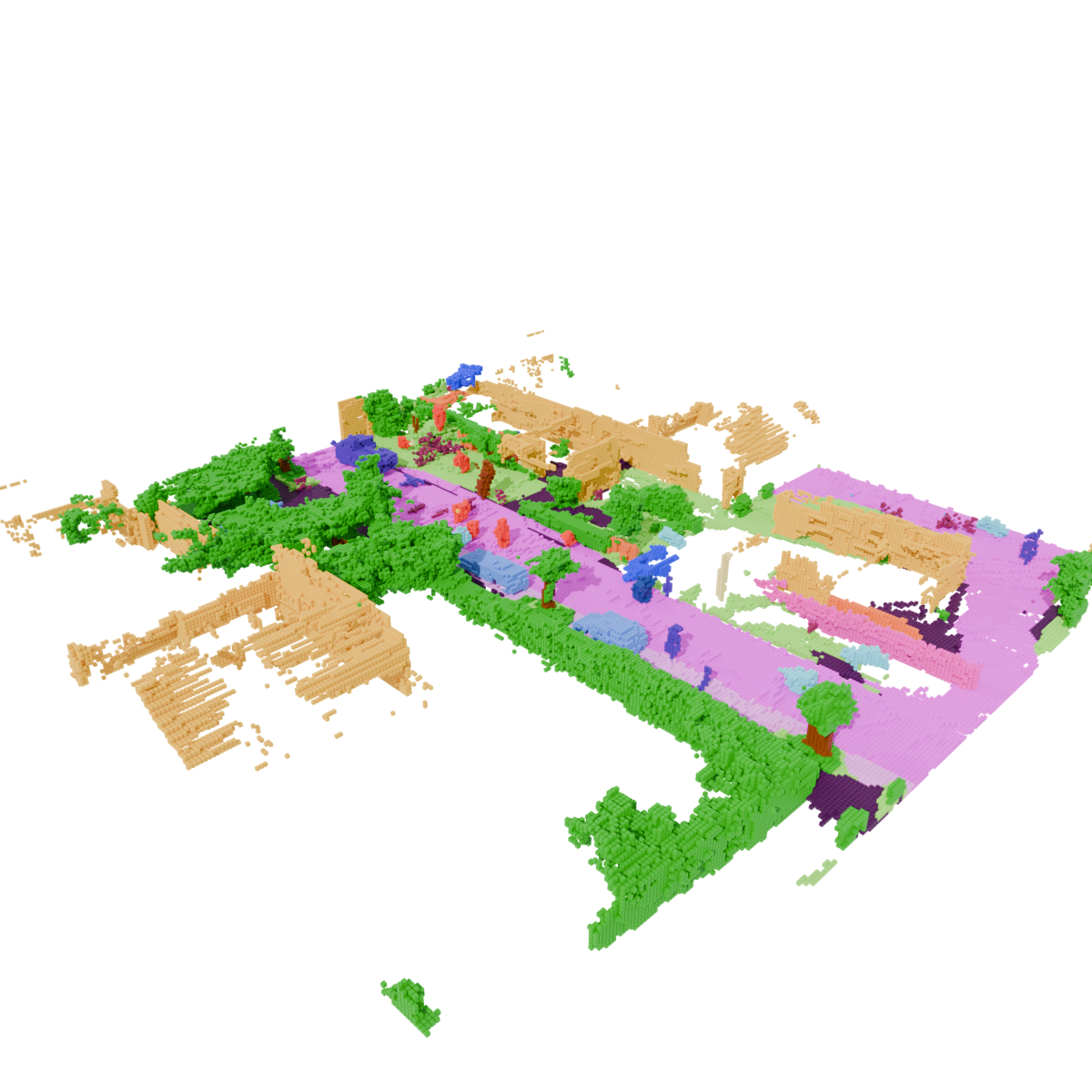}};
        \draw[flow] (s1.east) -- (s2.west) node[midway,above=1pt,badge]{diffuse};
        \draw[flow] (s2.east) -- (s3.west) node[midway,above=1pt,badge]{diffuse};
        \node[sub,below=1pt of s1] {(a)~$\mathcal{S}_1$: $32^2{\times}4$};
        \node[sub,below=1pt of s2] {(b)~$\mathcal{S}_2$: $64^2{\times}8$};
        \node[sub,below=1pt of s3] {(c)~$\mathcal{S}_3$: $256^2{\times}32$};

        \node[panel] (pb) at (2.35,0) {\includegraphics[width=\PW]{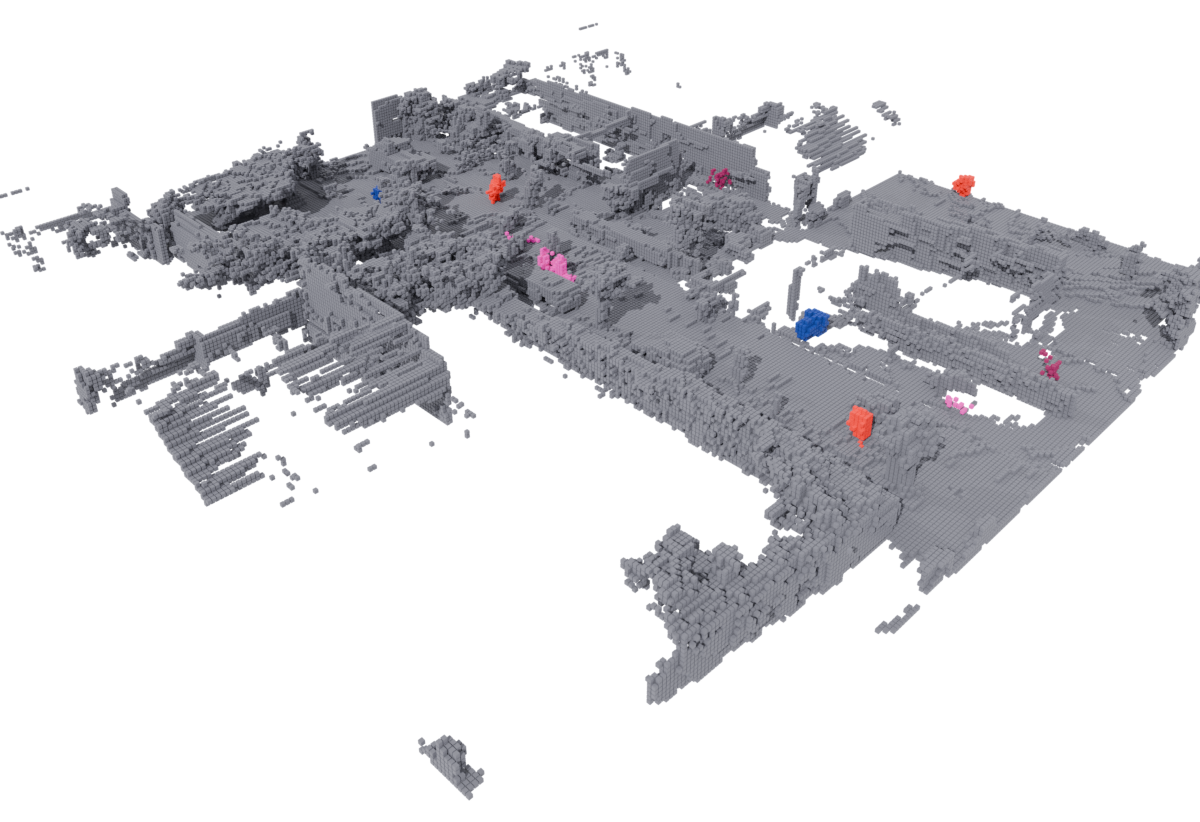}};
        \node[panel] (sp) at (9.55,0) {\includegraphics[width=\PW]{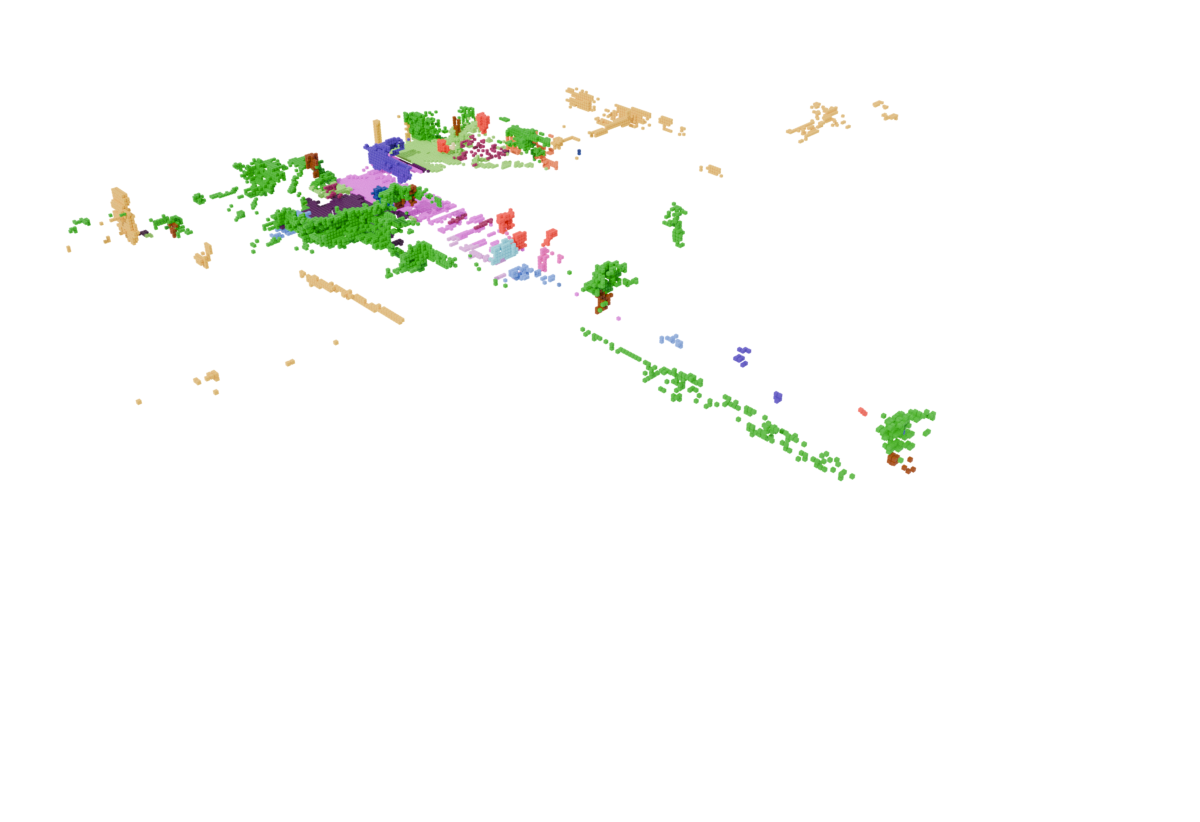}};
        \draw[flow] (pb.east) -- (sp.west) node[midway,above=1pt,badge]{\textsc{Halo}};
        \node[sub,below=1pt of pb] {(d) +\,object-bank rare classes $=\widetilde{\mathbf{Y}}$};
        \node[sub,below=1pt of sp] {(e) sparse sweep $\widetilde{\mathbf{X}}$};

        \coordinate (w1) at (15.2,5.15);
        \coordinate (w2) at (15.2,2.3);
        \coordinate (w3) at (2.35,2.3);
        \draw[flow] (s3.east) -- (w1) -- (w2) -- (w3) -- (pb.north);
        \node[badge] at (9.9,2.3) {$D_{\mathrm{JS}}$ screen};

        \draw[black!45,line width=0.7pt]
            ($(pb.south west)+(0,-1.05)$) -- ($(sp.south east)+(0,-1.05)$);
        \node[badge] at ($(pb.south)!0.5!(sp.south)+(0,-1.05)$)
            {paired training sample $(\widetilde{\mathbf{X}},\,\widetilde{\mathbf{Y}})$};
    \end{tikzpicture}}%
    \caption{\textbf{One PS\textsuperscript{3} pair.} Pyramid diffusion generates a scene ($\mathcal{S}_1\!\to\!\mathcal{S}_2\!\to\!\mathcal{S}_3$, a--c); the $D_{\mathrm{JS}}$-screened sample receives object-bank rare classes, giving $\widetilde{\mathbf{Y}}$ (d; pasted objects in colour), which \textsc{Halo} ray-traces into the sparse sweep $\widetilde{\mathbf{X}}$ (e).}
    \label{fig:da_visualizations}
\end{figure*}

A completion model learns only from \emph{paired} examples (a sparse sweep aligned to the complete, per-voxel-labelled scene it should recover), and outdoor pairs are expensive and long-tailed at once. PS\textsuperscript{3} manufactures them offline: it generates the dense scene $\widetilde{\mathbf{Y}}$, then derives the sparse observation $\widetilde{\mathbf{X}}$ that would have produced it (\cref{fig:ps3_pipeline}; \cref{fig:da_visualizations} follows one pair through it). Nothing in the pool depends on the network being trained, so it is reusable augmentation.

\paragraph{Scene generation} \looseness=-1 The full-resolution grid is high-dimensional and overwhelmingly empty, so we generate it coarse-to-fine with pyramid discrete diffusion~\cite{liu2024pyramid}. Three levels of the multinomial diffusion of \cref{sec:prelim} are each conditioned on the trilinearly upsampled scene from the level below, so coarse geometry guides fine detail. The coarsest level, lacking a coarser prior, trains on masked complete scenes instead, under the context patterns of supplementary Appendix~G.

\paragraph{Distribution filtering} Cascade sampling is fast but imperfect: some raw scenes are nearly empty, single-class, geometrically impossible, or distribution-shifted. Three inexpensive structural checks reject the first three: a plausible occupancy ratio, road plus one further structural class, and a gravity-consistent vertical layout. Each survivor is then screened on the Jensen--Shannon divergence between its class histogram and a corpus prior estimated on held-out real scenes, keeping scenes at $D_{\mathrm{JS}}\le\tau{=}0.35$ and then the lowest-$D_{\mathrm{JS}}$ half. The loose $\tau$ constrains class balance and leaves geometry to the structural checks.

\paragraph{Rare-class pasting} The generator reproduces the dominant layout faithfully but seldom places the safety-critical classes at all. We therefore graft real instances into each filtered scene, a 3D copy-paste from an offline bank of voxelised instances harvested by connected components over the real training labels. Per scene we visit the rare classes in a fixed order and raise each toward a minimum voxel count, under a paste budget shared across all of them, so the target is attempted rather than met. Every paste must fit the grid, overwrite no other class, and rest on a drivable surface rather than float.

\paragraph{Observation rendering} \looseness=-1 A generated scene enters training only alongside the sparse input a real sensor would return. Subsampling occupied voxels will not do: a rotating LiDAR samples through a fixed beam fan, dense near the sensor and ground, sparse far away and overhead. Our Hardware-Aware \mbox{LiDAR Observation} (\textsc{Halo}) operator marches the $64$ non-uniformly spaced HDL-64E beams~\cite{velodyne2011hdl64e} from a fixed origin through $\widetilde{\mathbf{Y}}$, so the sweep inherits a real scan's ring structure and range falloff. Two choices serve the long tail: dual returns per beam, as in full-waveform practice~\cite{winiwarter2022helios}, recover thin or occluded structures a first-hit tracer erases, and per-return range and angular jitter~\cite{manivasagam2020lidarsim} injects real measurement error. An occupied voxel no sample lands in is marked \texttt{invalid}. An empty one never is. \Cref{alg:halo} states it; sensor calibration, filter thresholds and paste parameters are in Appendix~A.

\begin{algorithm}[!t]
\caption{\textsc{Halo}: Hardware-Aware LiDAR Observation. $\textsc{Voxel}(\cdot)$ maps a metric point to its grid cell at edge $\Delta_v{=}0.2$\,m, or $\textsc{Null}$ outside the grid, so $\textsc{Voxel}(\mathbf{o}){=}(0,128,10)$. A beam at angles $(\phi,\psi)$ has direction $\mathbf{d}(\phi,\psi){=}(\cos\phi\cos\psi,\,\cos\phi\sin\psi,\,\sin\phi)$. Beam angles, the march step $\delta t$, post-hit skip $\Delta t$, range limit $R_{\max}$, $\textsc{MaxReturns}$, $\textsc{SamplesPerReturn}$ and the noise scales are tabulated in Appendix~A.}
\label{alg:halo}
\small
\begin{algorithmic}[1]
\Require Complete label grid $\widetilde{\mathbf{Y}}$ with $\textsc{Empty}{=}0$; sensor origin $\mathbf{o}$
\Ensure Sparse mask $\widetilde{\mathbf{X}}$; invalid mask $\mathbf{V}$ ($1{=}$ occupied, unobserved)
\State $\widetilde{\mathbf{X}} \gets \mathbf{0}$
\ForAll{beams $(\phi,\psi)$}
  \State $t \gets 0$;\; $r \gets 0$;\; $v_{\text{prev}} \gets \textsc{Null}$
  \While{$t < R_{\max}$ \textbf{and} $r < \textsc{MaxReturns}$}
    \State $t \gets t + \delta t$;\quad $v \gets \textsc{Voxel}(\mathbf{o} + t\,\mathbf{d}(\phi,\psi))$
    \State \textbf{if} $v = \textsc{Null}$ \textbf{then break}
    \State \textbf{if} $\widetilde{\mathbf{Y}}[v] = \textsc{Empty}$ \textbf{or} $v = v_{\text{prev}}$ \textbf{then continue}
    \For{$k = 1$ \textbf{to} $\textsc{SamplesPerReturn}$}
      \State $(\epsilon_r,\epsilon_\psi,\epsilon_\phi) \sim \mathcal{N}\bigl(\mathbf{0},\,\operatorname{diag}(\sigma_r^2,\sigma_\psi^2,\sigma_\phi^2)\bigr)$
      \State $u \gets \textsc{Voxel}\bigl(\mathbf{o} + (t{+}\epsilon_r)\,\mathbf{d}(\phi{+}\epsilon_\phi,\,\psi{+}\epsilon_\psi)\bigr)$
      \State \textbf{if} $u \neq \textsc{Null}$ \textbf{and} $\widetilde{\mathbf{Y}}[u] \neq \textsc{Empty}$ \textbf{then} $\widetilde{\mathbf{X}}[u] \gets 1$
    \EndFor
    \State $v_{\text{prev}} \gets v$;\; $r \gets r + 1$;\; $t \gets t + \Delta t$ \Comment{clear the hit cell}
  \EndWhile
\EndFor
\State $\mathbf{V} \gets [\widetilde{\mathbf{Y}} \neq \textsc{Empty}] \wedge \neg\widetilde{\mathbf{X}}$ \Comment{occupied, never sampled}
\State \Return $(\widetilde{\mathbf{X}}, \mathbf{V})$
\end{algorithmic}
\end{algorithm}

\subsection{Semantic-Guided Generative Scene Completion (SGSC)}
\label{sec:sgsc}

To complete a scene with no base prediction to start from, we introduce SGSC, which casts completion as conditional generation: a multinomial diffusion process carries per-voxel categorical distributions from a uniform noise prior to the complete semantic scene. What the chain reads is the whole of the design, a BEV semantic map and per-voxel segmentation features from a single sweep, so partial semantics reach it before it completes the geometry. No completion model enters the pipeline at any stage.

\paragraph{Denoiser and conditioning}

\looseness=-1 The ground-truth grid is embedded on the product simplex $\mathcal{M}$ of \cref{sec:prelim}. A chain initialised at $\mathbf{u}$ generates unconditionally, so the observation must steer every reverse step. From a single sweep $\mathbf{X}$ we derive two complementary conditioning streams $\mathbf{c} = \{\mathbf{B}, \mathbf{F}\}$. A learned sparse-3D BEV head $h_{\mathrm{bev}}$ gives a per-cell label map $\mathbf{B}\!\in\!\{0,\ldots,K{-}1\}^{L\times W}$, replicated along $Z$ and embedded by the same convolution as the noisy iterate. A frozen LSK3DNet~\cite{feng2024lsk3dnet} backbone yields a per-voxel feature volume $\mathbf{F}$ (any voxel-aligned LiDAR-segmentation backbone can replace it), read by an auxiliary sparse residual encoder over the ${\sim}1\%$ occupied voxels (architecture summary in Appendix~G). $h_{\mathrm{bev}}$ is trained separately and its label maps are precomputed once over the pool, so it is frozen with respect to $f_\theta$ and no gradient crosses the integer-valued $\mathbf{B}$. The two streams reach a level differently. $\mathcal{E}_\ell(\mathbf{B})$ embeds $\mathbf{B}$ once, as a $K$-channel one-hot volume through a $3{\times}3{\times}3$ convolution. It average-pools the volume by a factor of two, $\ell$ times, and projects it to level $\ell$'s width with a further $3{\times}3{\times}3$ convolution owned by that block. $\mathcal{E}_\ell(\mathbf{F})$ is that same per-block projection applied to the sparse encoder's own level-$\ell$ output, which needs no pooling. Both are injected \emph{additively},
\begin{equation}
    \mathbf{h}^{(\ell)} \leftarrow \mathbf{h}^{(\ell)} + \mathcal{E}_\ell(\mathbf{B}) + \mathcal{E}_\ell(\mathbf{F}),
    \label{eq:sgsc_inject}
\end{equation}
not by FiLM's~\cite{perez2018film} feature-wise affine modulation; the additive form keeps observation and iterate in one feature space. Neither stream is a completed scene: SGSC \emph{generates} the volume from $\mathbf{c}$ rather than refining an existing prediction.

\begin{figure*}[!tbp]
    \centering
    \includegraphics[width=\textwidth]{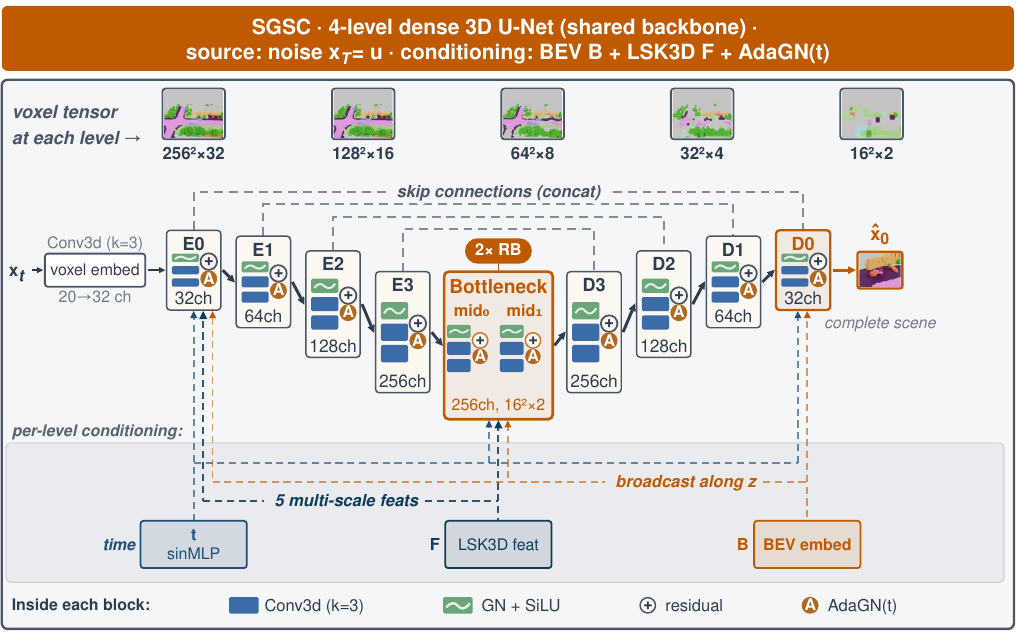}
    \caption{\textbf{The SGSC denoiser in its generation role.} A four-level dense 3D U-Net (channels $32\!\to\!256$) generates the scene from the uniform prior $\mathbf{x}_T{=}\mathbf{u}$, conditioned on a BEV map $\mathbf{B}$ and per-voxel LSK3DNet features $\mathbf{F}$ from one sweep, injected additively at every level (\cref{eq:sgsc_inject}); the timestep enters only through adaptive group normalisation (AdaGN), and the bottleneck stacks two residual blocks (RB). The same architecture, retrained against a structured source, drives S\textsuperscript{2}D\textsuperscript{2} refinement (\cref{fig:s2d2}).}
    \label{fig:sgsc}
\end{figure*}

The reverse process is driven by a single denoiser $f_\theta$ (\cref{fig:sgsc}): a four-level dense 3D U-Net (channels $32\!\to\!256$) with a transposed-convolution decoder and encoder skips, in the standard diffusion design~\cite{ho2020denoising,dhariwal2021diffusion}. The timestep enters only through adaptive group normalisation (AdaGN)~\cite{dhariwal2021diffusion},
\begin{equation}
    \mathrm{AdaGN}(\mathbf{h}, t) = \mathbf{s}(t)\odot \mathrm{GN}(\mathbf{h}) + \mathbf{b}(t),
    \qquad (\mathbf{s}, \mathbf{b}) = g\bigl(\boldsymbol{\tau}(t)\bigr),
    \label{eq:sgsc_film}
\end{equation}
with $g$ a linear projection of the sinusoidal embedding $\boldsymbol{\tau}(t)$; the observation instead enters additively (\cref{eq:sgsc_inject}). A final convolution gives per-voxel logits whose softmax is the $\mathbf{x}_0$-prediction $\hat{p}_\theta(\cdot\mid\mathbf{x}_t, t, \mathbf{c}) = \operatorname{softmax}\!\bigl(f_\theta(\mathbf{x}_t, t, \mathbf{c})\bigr)$ used by the reverse step. Per-level widths are in Appendix~G. The denoiser totals ${\sim}35$M parameters.

\paragraph{Training objective}
Training minimises the per-step KL between the analytic posterior (\cref{eq:sgsc_posterior}) and the parametric reverse (\cref{eq:sgsc_reverse}),
\begin{equation}
\begin{aligned}
    \mathcal{L}_{\mathrm{SGSC}} = \mathbb{E}_{t,\,\mathbf{x}_0,\,\mathbf{x}_t \sim q(\mathbf{x}_t \mid \mathbf{x}_0),\,\mathbf{c}} \bigl[\, & D_{\mathrm{KL}}\bigl( q(\mathbf{x}_{t-1} \mid \mathbf{x}_t, \mathbf{x}_0) \;\big\|\\
    &\ \, p_\theta(\mathbf{x}_{t-1} \mid \mathbf{x}_t, \mathbf{c}) \bigr) \bigr],
\end{aligned}
    \label{eq:sgsc_loss}
\end{equation}
with $t$ uniform on $\{1,\ldots,T\}$; summed over $t$ this is the standard variational upper bound on $-\log p_\theta(\mathbf{x}_0\!\mid\!\mathbf{c})$~\cite{hoogeboom2021argmax}; its parameter-free terminal term drops out, its $t{=}1$ term is cross-entropy against the one-hot target. We augment the KL with Lov\'asz-softmax~\cite{berman2018lovasz} and a bottleneck-auxiliary surrogate (weights in Appendix~G).

\paragraph{Full posterior sampling at inference}
Sampling starts from the uniform prior $\tilde{\mathbf{x}}_T = \mathbf{u}$, the tilde marking a sampler iterate, and runs the parametric reverse~\cref{eq:sgsc_reverse} from $t{=}T$ to $1$ under $f_{\theta'}$, the exponential-moving-average (EMA) copy of $\theta$ ($T{=}100$, EMA decay $0.9999$). It draws each $\tilde{\mathbf{x}}_{t-1}\!\sim\!p_{\theta'}(\tilde{\mathbf{x}}_{t-1}\!\mid\!\tilde{\mathbf{x}}_t,\mathbf{c})$ and reads off $\hat{\mathbf{Y}}_v = \argmax_k \tilde{\mathbf{x}}_0[v,k]$ from the $t{=}0$ draw (full loop in Appendix~G). Prediction is stochastic and we fix the seed. Unlike the refinement regime below, it cannot be shortened: one-hot targets lie near the maximal simplex separation from $\mathbf{u}$, so transport spans the whole trajectory and from-noise generation costs $T$ denoiser evaluations per scene (few-step distillation is future work).

\begin{figure*}[!tbp]
    \centering
    \includegraphics[width=0.82\textwidth]{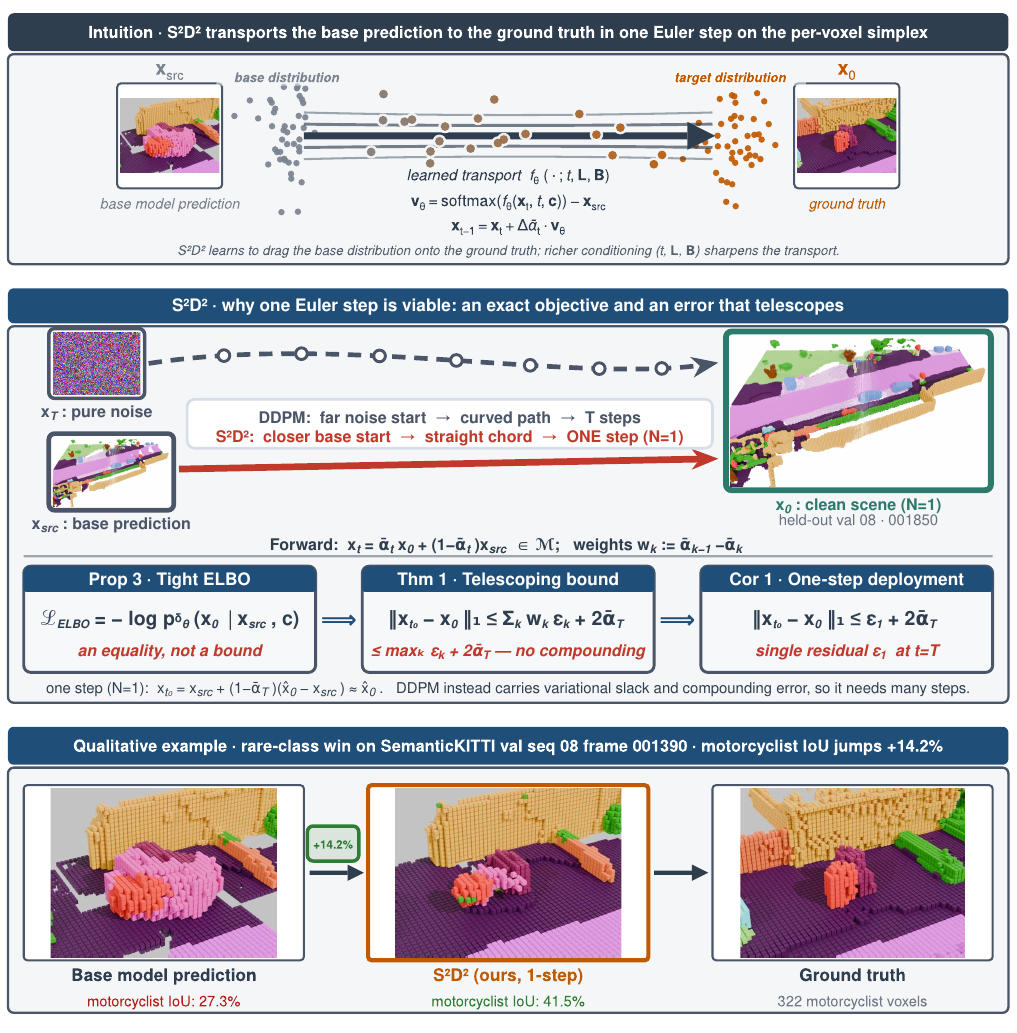}
    \caption{\textbf{S\textsuperscript{2}D\textsuperscript{2} refinement: the shared denoiser's second role.} The velocity field carries the base prediction $\mathbf{x}_{\mathrm{src}}$ to ground truth $\mathbf{x}_0$ in one Euler step on the per-voxel simplex (top); the identity and error bound rule out the amplification that would force the full chain (middle; Prop.~3, Thm.~1 and Cor.~1 of Appendix~B); a motorcyclist scene, frozen base vs.\ base$+$S\textsuperscript{2}D\textsuperscript{2} at $N{=}1$, one denoiser evaluation (bottom).}
    \label{fig:s2d2}
\end{figure*}

\subsection{Structured Source Discrete Diffusion (\texorpdfstring{S\textsuperscript{2}D\textsuperscript{2}}{S2D2})}
\label{sec:s2d2}

To improve a completion without retraining the model that produced it, we introduce S\textsuperscript{2}D\textsuperscript{2}, which refines a frozen model's prediction by flow matching on the per-voxel label simplex (\cref{fig:s2d2}). Standard flows start from a Gaussian or uniform prior and so discard whatever the base already recovered. We replace that prior with a \emph{structured} source, the one-hot per-voxel \emph{argmax} prediction $\mathbf{x}_{\mathrm{src}}$ of a frozen base $g_{\mathrm{base}}$ (\emph{e.g.}, SCPNet~\cite{xia2023scpnet}), keeping the one-hot ground truth $\mathbf{x}_0$ as the target. Both endpoints are then fixed and the path between them is straight, which is what admits a single deployment step and leaves the base untouched, its logits included.

\paragraph{Forward process from a structured source}
We instantiate every ingredient on the product simplex $\mathcal{M}=\prod_v\Delta^{K-1}$ of \cref{sec:prelim}. The source is the base's decoded label map rather than its logits, which fixes what each uplift is measured against; any simplex-valued source would serve, and the confidence the argmax discards is an unevaluated extension. With both endpoints fixed we read the interpolant on the diffusion clock $\bar{\alpha}_t = \prod_{r=1}^{t}(1-\beta_r)$, the cumulative-retention schedule of \cref{sec:prelim} taken as a deterministic flow time: $\bar{\alpha}_0 = 1$ is the target end ($\mathbf{z}_1 \leftrightarrow \mathbf{x}_0$), $\bar{\alpha}_T \approx 0$ the source end ($\mathbf{z}_0 \leftrightarrow \mathbf{x}_{\mathrm{src}}$). The interpolant is $\mathbf{z}_s$ at $s = \bar{\alpha}_t$ (the implementation takes $\beta_t$ linear from $10^{-4}$ to $0.1$ over $T{=}100$ steps, so $\bar{\alpha}_T \approx 5.6\times10^{-3}$):
\begin{equation}
    \mathbf{x}_t \;=\; \bar{\alpha}_t \, \mathbf{x}_0 \;+\; (1 - \bar{\alpha}_t) \, \mathbf{x}_{\mathrm{src}}, \qquad t \in \{0, 1, \ldots, T\}.
    \label{eq:forward}
\end{equation}
The forward kernel is therefore a Dirac on \cref{eq:forward}, not the stochastic Markov corruption of a D3PM~\cite{austin2021structured}. Each $\mathbf{x}_t$ is a convex combination of two simplex points, so the path stays in $\mathcal{M}$ with no mass reallocation and no transition matrix $\mathbf{Q}_t$ (\emph{simplex invariance}), and its terminal $\mathbf{x}_T \approx \mathbf{x}_{\mathrm{src}}$ keeps it anchored to the base's estimate.

\Cref{eq:forward} is the \emph{mean path}, and the two ends instantiate it differently: training feeds per-voxel one-hot draws $\check{\mathbf{x}}_t\sim\mathrm{Cat}(\mathbf{x}_t)$, whose mean is exactly $\mathbf{x}_t$, while the sampler never draws and runs on the path points themselves. Keeping the training state categorical reuses the multinomial-diffusion posterior with the source in place of the uniform prior, and keeps every per-step term a finite categorical KL. Throughout, $\mathbf{x}_t$ is the path point, $\check{\mathbf{x}}_t$ the categorical draw training takes from it, $\tilde{\mathbf{x}}_t$ the sampler's own iterate, and $\delta$ marks objects evaluated on the path point.

\paragraph{Source conditioning}
The denoiser is conditioned on a pair of streams derived from the base prediction and raw sweep, $\mathbf{c}=\{\mathbf{B},\mathbf{L}\}$. $\mathbf{B}$ is the \emph{base-derived} BEV semantic map, the parameter-free topmost-non-empty-class projection of the base's 3D prediction, and $\mathbf{L}$ a binary occupancy grid voxelised from the raw scan. $f_\theta$ reads a state $\mathbf{z}$ (the path point here, the categorical draw in training, the sampler iterate at inference) and emits the clean-scene estimate $\hat{\mathbf{x}}_0(\mathbf{z}) := \operatorname{softmax}\bigl(f_\theta(\mathbf{z},t,\mathbf{c})/\tau_{\mathrm{smp}}\bigr)$. The sampling temperature is $\tau_{\mathrm{smp}}{=}1$ in every main-body result, immaterial at the deployed single step (\cref{sec:ema_ablation}). We drop arguments when clear. The BEV stream is left uncorrected: refining the map in 2D leaves the SSC output unchanged to within run-to-run variation (\cref{sec:bev_ablation}).

The occupancy grid is the only input whose regime differs between training and deployment. It is multi-frame-accumulated at training, following SCPNet's multi-frame teacher~\cite{xia2023scpnet}, and single-frame at deployment. The shift is benign: retraining end-to-end with a single-frame grid at both ends costs a fraction of a point (\cref{sec:single_frame_retrain_ablation}), so we keep the accumulated grid at training and deploy single-frame.

S\textsuperscript{2}D\textsuperscript{2} reuses the denoiser \emph{architecture} of \cref{sec:sgsc}, the four-level dense 3D U-Net of \cref{fig:sgsc}, but not its weights: $f_\theta$ is trained from scratch against the structured source. The grid $\mathbf{L}$ enters through the auxiliary sparse encoder that processed $\mathbf{F}$ in \cref{sec:sgsc}: the structured source already carries dense per-voxel semantics, so the conditioning need only reinject the scan's geometric evidence. Both enter additively at every level as in \cref{eq:sgsc_inject}, with $\mathcal{E}_\ell(\mathbf{L})$ in place of $\mathcal{E}_\ell(\mathbf{F})$.

\paragraph{Prediction target and the per-step objective}
The displacement to be estimated along this path is the constant $\mathbf{x}_0-\mathbf{x}_{\mathrm{src}}$, so the velocity the network realises through $\hat{\mathbf{x}}_0$ is
\begin{equation}
    \mathbf{v}_\theta(\mathbf{x}_t,t,\mathbf{c};\mathbf{x}_{\mathrm{src}}) \;=\; \hat{\mathbf{x}}_0 - \mathbf{x}_{\mathrm{src}}.
    \label{eq:s2d2_velocity}
\end{equation}
The estimate $\hat{\mathbf{x}}_0$ is the network's only prediction target, shared by training and sampling (also the SGSC denoiser's $\mathbf{x}_0$-prediction) and free to lie in the simplex interior; velocities are read in flow time $s=\bar{\alpha}_t$. The source is a known constant of the sample. It enters the posterior and the plug-in reverse algebraically, and the denoiser through a learned convolution summed into the input features, alongside the state and BEV map $\mathbf{B}$. It also shrinks the regressed displacement: $\mathbf{x}_{\mathrm{src}}=\mathbf{u}$ gives the from-noise regime's near-maximal one, closed only over many steps; the structured source leaves one supported only on the base's erroneous voxels (small at high top-1 accuracy). Short and, by the straight interpolant, constant along the path, it needs one Euler step to traverse, up to the endpoint residual (telescoping error bound, Appendix~B).

The objective is the discrete-diffusion variational one rather than a plain velocity regression, which under \cref{eq:s2d2_velocity} would reduce to an $L_2$ fit of $\hat{\mathbf{x}}_0$ onto $\mathbf{x}_0$: on the deterministic path the variational bound is not a bound but an exact conditional likelihood (Appendix~B). Training tuples are formed offline from the frozen base's argmax predictions on the training frames. Training instantiates the mean path stochastically through the draws $\check{\mathbf{x}}_t \sim \mathrm{Cat}(\mathbf{x}_t)$ introduced above; the per-step target is the multinomial-diffusion posterior \emph{expression}~\cite{hoogeboom2021argmax} with the source in place of the uniform prior,
\begin{equation}
\begin{aligned}
    q(\check{\mathbf{x}}_{t-1}\!\mid\!\check{\mathbf{x}}_t,\mathbf{x}_0,\mathbf{x}_{\mathrm{src}}) \;\propto\;
      & \bigl[\alpha_t\,\check{\mathbf{x}}_t + (1{-}\alpha_t)\,\mathbf{x}_{\mathrm{src}}\bigr] \odot \\
      & \quad \bigl[\bar{\alpha}_{t-1}\,\mathbf{x}_0 + (1{-}\bar{\alpha}_{t-1})\,\mathbf{x}_{\mathrm{src}}\bigr],
\end{aligned}
    \label{eq:s2d2_posterior}
\end{equation}
\looseness=-1 with $\alpha_t = 1-\beta_t$ and $\odot$ the elementwise product. Setting $\mathbf{x}_{\mathrm{src}} = \mathbf{u}$ recovers \cref{eq:sgsc_posterior}; otherwise \cref{eq:s2d2_posterior} is a construction, not a Bayes identity. This is the form the released trainer evaluates.

The reverse reuses the endpoint estimate in the same plug-in form as \cref{eq:sgsc_reverse}, $p_\theta(\check{\mathbf{x}}_{t-1}\!\mid\!\check{\mathbf{x}}_t,\mathbf{x}_{\mathrm{src}},\mathbf{c}) := q(\check{\mathbf{x}}_{t-1}\!\mid\!\check{\mathbf{x}}_t,\hat{\mathbf{x}}_0,\mathbf{x}_{\mathrm{src}})$: the two regimes share one reverse parameterisation and differ only in the noise target: uniform $\mathbf{u}$ there, the structured source $\mathbf{x}_{\mathrm{src}}$ here. Although a one-hot source forfeits the uniform prior's strict positivity, the KL stays finite ($\operatorname{supp} q \subseteq \operatorname{supp} p_\theta$; support argument accompanying the training algorithm in Appendix~G). At $t{=}1$ the second bracket collapses to $\mathbf{x}_0$ and we take the usual reconstruction cross-entropy against the label~\cite{ho2020denoising,hoogeboom2021argmax}. With $t$ uniform on $\{1,\ldots,T\}$ and the expectation including the draw of $\check{\mathbf{x}}_t$, the objective is
\begin{equation}
\begin{aligned}
    \mathcal{L}_{\mathrm{KL}} = \mathbb{E}_{t,\,\check{\mathbf{x}}_t,\,(\mathbf{x}_0, \mathbf{x}_{\mathrm{src}}, \mathbf{c})} \bigl[ \;
      & D_{\mathrm{KL}} \bigl( q(\check{\mathbf{x}}_{t-1} \mid \check{\mathbf{x}}_t, \mathbf{x}_0, \mathbf{x}_{\mathrm{src}}) \\
      & \;\;\bigl\|\; p_\theta(\check{\mathbf{x}}_{t-1} \mid \check{\mathbf{x}}_t, \mathbf{x}_{\mathrm{src}}, \mathbf{c}) \bigr) \bigr].
\end{aligned}
    \label{eq:kl_loss}
\end{equation}

\paragraph{How exact the objective is, and where it stops}
On the deterministic mean path of \cref{eq:forward}, per-step matching admits an exact
conditional-likelihood identity. Started from the source-conditional prior
$\delta_{\mathbf{x}_{\mathrm{src}}}$, under a strictly decreasing, exactly terminating schedule and
one-hot targets, the usual variational bound collapses to an \emph{equality}. This holds because the
Dirac kernel makes the Jensen step exact and the source-conditional prior matches the forward
terminal. The identity transfers to the trained objective of \cref{eq:kl_loss} only approximately:
that objective scores the one-hot draws $\check{\mathbf{x}}_t$, not the mean-path points the identity
governs. The two share a minimiser at $\hat{\mathbf{x}}_0 = \mathbf{x}_0$, so the transfer is exact
in the realisable limit alone. Statement and proof are in Appendix~B.

The shipped schedule does not terminate exactly, and the consequence is confined to deployment: with $\bar{\alpha}_T > 0$ the forward terminal sits just short of $\mathbf{x}_{\mathrm{src}}$, so we do not claim the deterministic-path bound there, and the effect is bounded by the $2\bar{\alpha}_T$ term below.

The deployed objective adds three task-specific terms to a per-voxel-weighted KL:
\begin{equation}
    \mathcal{L} = \tilde{\mathcal{L}}_{\mathrm{KL}} + \lambda_{\mathrm{Lov}}\bigl(\mathcal{L}_{\mathrm{Lov}}^{\mathrm{all}} + \mathcal{L}_{\mathrm{Lov}}^{\mathrm{occ}}\bigr) + \lambda_a \, \mathcal{L}_{\mathrm{aux}},
    \label{eq:full_loss}
\end{equation}
\looseness=-1 where $\tilde{\mathcal{L}}_{\mathrm{KL}}$ is \cref{eq:kl_loss} with per-voxel weights. The two Lov\'{a}sz-softmax terms~\cite{berman2018lovasz} share the weight $\lambda_{\mathrm{Lov}}$ and apply the same IoU surrogate to $\hat{\mathbf{x}}_0$ over different voxel populations. $\mathcal{L}_{\mathrm{Lov}}^{\mathrm{all}}$ runs over the whole grid, so it prices predicted mass on empty space; $\mathcal{L}_{\mathrm{Lov}}^{\mathrm{occ}}$ runs over occupied voxels only, so it prices confusion among the classes. $\mathcal{L}_{\mathrm{aux}}$ is a cross-entropy of $\hat{\mathbf{x}}_0$ against the clean labels, scaled by $(1-t/T)+1$, and added \emph{alongside} the per-step KL rather than substituted for it. The KL term is weighted by focal~\cite{lin2017focal}, effective-number class-balanced~\cite{cui2019class}, and observation-aware terms (the last tripling the weight of voxels the sweep returns). These steer the signal onto the occupied voxels ($\sim$$5\%$ of the grid, yet the ones scored by per-class IoU) and, within them, the rare classes. The coefficients $\lambda_{\mathrm{Lov}},\lambda_a$, the focal exponent, and the class and observation weights are tabulated in Appendix~G, which also traces a training step end-to-end.

\paragraph{Inference by correction sampling}
The sampler is initialised at the structured source, not at noise: at $t{=}T$ the training draw $\check{\mathbf{x}}_T$ is an $O(\bar{\alpha}_T)$ perturbation of $\mathbf{x}_{\mathrm{src}}$, identical wherever the base is right.

\looseness=-1 Every update then moves along one direction. The network estimates the path's \emph{endpoint} $\hat{\mathbf{x}}_0$ at every $t$ and \cref{eq:forward} is straight, so the displacement left to travel is always $\hat{\mathbf{x}}_0 - \mathbf{x}_{\mathrm{src}}$ up to a schedule factor, in place of the per-step slice a standard reverse sampler follows:
\begin{equation}
    \tilde{\mathbf{x}}_{t-1} = \tilde{\mathbf{x}}_t + (\bar{\alpha}_{t-1} - \bar{\alpha}_t) \cdot \big( \hat{\mathbf{x}}_0 - \mathbf{x}_{\mathrm{src}} \big).
    \label{eq:correction_step}
\end{equation}
In flow time $s=\bar{\alpha}_t$ this is the explicit Euler step for the velocity of \cref{eq:s2d2_velocity}, and Cold Diffusion's improved sampler~\cite{bansal2022cold}, exact for $\mathbf{x}_0$-affine degradations such as ours. Iterated from $\tilde{\mathbf{x}}_{t_N} = \mathbf{x}_{\mathrm{src}}$ over an $N$-step schedule $T = t_N \ge \cdots \ge t_0 = 0$ ($N$ counts denoiser evaluations), it carries the iterate toward $\mathbf{x}_0$, and the label is $\hat{\mathbf{Y}}_v = \argmax_k \tilde{\mathbf{x}}_{t_0}[v,k]$ (full loop in Appendix~G).


\paragraph{How the error behaves over the chain}
Because the path is linear, summing \cref{eq:correction_step} along the schedule telescopes, and the
final iterate's deviation from $\mathbf{x}_0$ is bounded by the largest per-step prediction residual
$\varepsilon_i := \lVert\operatorname{softmax}(f_\theta(\tilde{\mathbf{x}}_{t_i},t_i,\mathbf{c})) -
\mathbf{x}_0\rVert_1$ plus a $2\bar{\alpha}_T$ offset from initialising at
$\mathbf{x}_{\mathrm{src}}$: $\lVert \tilde{\mathbf{x}}_{t_0} - \mathbf{x}_0\rVert_1 \le
\max_i\varepsilon_i + 2\bar{\alpha}_T$. The offset is the cost of the implementation's nonzero
terminal rather than of the method. Iterated correction therefore does not amplify per-step error,
with no Lipschitz assumption, and deployment collapses the sum to one term. At $N{=}1$ the error is
governed by the \emph{single} residual at $t{=}T$: on the discrete simplex the counterpart of
InDI's single-step regime~\cite{delbracio2023inversion}. There one step minimises distortion at the cost of perceptual quality---the trade a voxel-labelling metric wants. Appendix~B
proves the bound, gives the conditions under which it is attained, and locates the irreducible
refiner error floor.

\looseness=-1 The bound is an $\ell_1$ distance while mIoU scores the \emph{argmax}. One line of algebra ties
the two: a voxel decodes correctly once its endpoint estimate clears the margin $\tfrac{1}{2} +
O(\bar{\alpha}_T)$, which the shipped schedule puts at $0.503$ (corollaries in Appendix~B). The theory therefore reaches the reported metric, but only so far: it
certifies the voxels that clear the margin and says nothing about the rest, where the empirical gap
between bases lives; it does not predict an mIoU.

\looseness=-1 Setting $\mathbf{x}_{\mathrm{src}}=\mathbf{u}$ recovers the from-noise chain of \cref{sec:sgsc}, and Appendix~B places \cref{eq:forward} beside its neighbours (D3PM~\cite{austin2021structured}, DDPM/DDIM~\cite{ho2020denoising,song2021denoising}, Rectified Flow~\cite{liu2023flow}, InDI~\cite{delbracio2023inversion} and Cold Diffusion~\cite{bansal2022cold}), marking which are literal instances.

\subsection{Why the Correction Works}
\label{sec:why_one_step}

Per-step matching supplies the usable signal. Where the base is right, $\mathbf{x}_{\mathrm{src}}=\mathbf{x}_0$ pins the path and the term vanishes for \emph{any} prediction; where it is wrong, \cref{eq:kl_loss} contrasts label against source rather than regressing the label alone. Replacing it with a timestep-uniform cross-entropy collapses training (\cref{sec:loss_ablation}).

The added terms tighten the objective rather than tilt it: all three score the same endpoint estimate and vanish as $\hat{\mathbf{x}}_0 \to \mathbf{x}_0$, so $\hat{\mathbf{x}}_0 = \mathbf{x}_0$ stays a minimiser of the composite. They carry the gradient wherever the KL term is flat because the base is already correct.

At $N{=}1$ the sampler starts at $\tilde{\mathbf{x}}_{t_N}=\mathbf{x}_{\mathrm{src}}$, so $\varepsilon_1$ is the residual of a single evaluation at an input the network was trained on, and the uniform-$t$ schedule is a curriculum for it. The telescoping bound does not assert that one step \emph{suffices}, since $\varepsilon_1$ is a network error it does not bound. What it rules out is the amplification that would otherwise force many, and that one step is enough is empirical (\cref{sec:step_reduction}).

\section{Experiments}
\label{sec:experiments}
\label{sec:setup}
\label{sec:hyperparams}

\noindent\textit{Datasets:} SemanticKITTI~\cite{behley2019semantickitti} is the standard outdoor SSC benchmark: $19{,}130$ training frames (seq.\ $00$--$07$, $09$--$10$), $4{,}071$ validation frames (seq.\ $08$, stride-1 voxel release) and $3{,}901$ hidden-test frames (seq.\ $11$--$21$, subsampled). Its target is a $256{\times}256{\times}32$ grid spanning $51.2{\times}51.2{\times}6.4$\,m at $0.2$\,m, labelled with $19$ classes plus empty/unlabelled, and its distribution is severely skewed: empty voxels are ${\sim}95\%$ of ground truth and motorcyclist $0.0037\%$ of labelled points (\cref{fig:teaser}c). PS\textsuperscript{3} (\cref{sec:data_aug}) adds $32{,}039$ synthetic pairs to these $19{,}130$ real frames, a $2.67\times$ expansion to $51{,}169$. Every part of it (the generator, the held-out scenes its distribution filter screens against, and the object bank) derives from this training split, and it augments SemanticKITTI training only. Two further datasets serve as zero-shot domains only, evaluated with the frozen SemanticKITTI checkpoint and no target labels or fine-tuning. They are SSCBench-KITTI360~\cite{li2024sscbench} (val seq.~$06$, $1{,}812$ frames, $16$ shared classes) and SemanticPOSS~\cite{pan2020semanticposs} (val seq.~$02$, ${\approx}500$ frames, the $11$-class TALoS Tab.~4 map, with ground truth reconstructed by us rather than official).

\noindent\textit{Metrics:} Following~\cite{roldao20223d,xia2023scpnet}, the primary metric is mIoU over the $19$ evaluation classes together with class-agnostic completion IoU (Comp.); precision and recall are quoted only where a diagnosis needs them. Every 3D number we report as a result comes from the official \texttt{semantic-kitti-api}, which excludes the \texttt{invalid} mask and so scores only voxels some sweep of the sequence reached. The never-observed volume, where a generative prior is most free to invent, lies outside every 3D number we report. Appendix~E bounds how much of our predicted occupancy falls there. Every mIoU we report as a result is a deployable single-frame measurement. The $2$D figures of \cref{sec:transfer} come from a separate BEV evaluator. All $\Delta$ values are computed from unrounded scores. Because mIoU weights every class alike and so hides the rare road users that safety depends on, we add three metrics that price the rare classes, two of them restricted to the \emph{vulnerable road user} (VRU) classes: person, bicyclist and motorcyclist. \textbf{Safety-Critical mIoU} (SC-mIoU) instead reweights all $19$ by four hazard tiers of our own choosing. ISO~21448~\cite{iso21448} assigns severity to a hazardous event in an operational situation rather than to a semantic class, so neither the tiers nor their order follows from a standard. Across the refined bases the ordering they produce is mIoU's at every weight vector we swept, so for those rows the SC-m column of \cref{tab:main_results,tab:portable_s2d2} is a robustness check rather than evidence separating methods; it is not ordering-free in general (\cref{sec:limits}). The tiers and the weight sweep are in Appendix~E. \textbf{VRU-IoU} is the hyperparameter-free mean IoU over the three VRU classes, and it is the tail metric that does reorder methods. \looseness=-1 \textbf{Detection-Window IoU} (DW-IoU) discounts per-class IoU by how many frames a system can deliver inside a reaction window,
\begin{equation}
    \mathrm{DW\text{-}IoU}_c = 1 - (1 - \mathrm{IoU}_c)^{R \cdot T_w},
    \label{eq:dwiou}
\end{equation}
with $R$ each system's deployed end-to-end rate and $T_w$ the window, fixed at $1.0$\,s; DW-VRU-IoU averages it over the three VRU classes. Because consecutive single-frame predictions are temporally correlated, DW-IoU is an optimistic upper bound on the benefit of a faster refresh, not a detection probability.

\noindent\textit{Training:} All runs use NVIDIA H100 $80$\,GB GPUs. SGSC ($\S$\ref{sec:sgsc}) and S\textsuperscript{2}D\textsuperscript{2} ($\S$\ref{sec:s2d2}) share one ${\sim}35$M-parameter denoiser architecture and one recipe: AdamW at lr~$10^{-4}$, effective batch~$4$ on $2{\times}$H100, $T{=}100$, and EMA weights. The headline frozen-base checkpoint is the best-EMA model at $40$K steps (${\sim}37$ GPU-hours) deployed at $N{=}1$. The alt-base and SGSC runs train to $100$K steps, and SGSC deploys under full posterior sampling ($N{=}T{=}100$) with conditioning $\mathbf{c}=\{\mathbf{B},\mathbf{F}\}$, the LiDAR-only BEV map and per-voxel LSK3D~\cite{feng2024lsk3dnet} features. The software stack, the seed, the noise schedule and the loss weights are in Appendix~G.

\noindent\textit{Baselines and protocol:} \looseness=-1 Published baselines are quoted at each method's own \emph{test}-leaderboard result, whereas our per-class and ablation evidence is measured on val seq.~$08$; \cref{tab:main_results} therefore carries an explicit \textbf{Eval} column and no row is read across splits. S3CNet~\cite{cheng2021s3cnet} enters the table on an uneven profile (completion IoU $45.6$ against $56.1$--$68.5$ for every other test row at mIoU $29.5\%$). Yet among single-sweep rows it leads every VRU class and takes that column, a gap on the safety metric we introduce that we do not close (\cref{sec:limits}; its per-class row in Appendix~D). DiffSSC predicts a point cloud and submitted it voxelised, and publishes no per-class breakdown. TALoS adapts the same base we refine~\cite{jang2024talos}, so its test row and ours start from one source. All numbers use a single seed without error bars, matching SSC convention; what reproduces under it is the per-class delta over the frozen base rather than the absolute mIoU, for the reason given with \cref{tab:perclass_delta}.

\subsection{Main Results}
\label{sec:main_results}
\label{sec:perclass}
\label{sec:qualitative_analysis}

\begin{table}[!t]
    \centering
    \scriptsize
    \setlength{\tabcolsep}{3pt}
    \caption{\textbf{Main results on SemanticKITTI.} Published baselines at their own \emph{test} leaderboard; our rows on both splits. S3CNet, not ours, takes the vulnerable road user (VRU) column (\cref{sec:limits}).}
    \label{tab:main_results}%
    \begin{tabular}{@{}llcccc@{}}
        \toprule
        Method & Eval & mIoU & Comp & SC-m & VRU \\
        \midrule
        LMSCNet~\cite{roldao2020lightweight}        & test & 17.6 & 56.7 & 12.4 & 0.0 \\
        SSA-SC~\cite{yang2021ssasc}                 & test & 23.5 & 58.8 & 17.8 & 2.6 \\
        JS3C-Net~\cite{yan2021sparse}               & test & 23.8 & 56.6 & 18.7 & 4.5 \\
        DiffSSC~\cite{cao2024diffssc}$^{\P}$           & test & 27.4 & \textbf{63.4} & \na & \na \\
        S3CNet~\cite{cheng2021s3cnet}                  & test & 29.5 & 45.6 & 29.9 & \textbf{32.6} \\
        SCPNet~\cite{xia2023scpnet} (published)     & test & 36.7 & 56.1 & 32.5 & 18.2 \\
        TALoS~\cite{jang2024talos}$^{\parallel}$      & test & 37.9 & 60.2 & 33.6 & 20.0 \\
        $\bullet$\,SCPNet\,$+$\,S\textsuperscript{2}D\textsuperscript{2} ($N{=}1$) & test & \textbf{38.8} & 58.9 & \textbf{34.7} & 21.6 \\
        $\bullet$\,\;\;{\textit{with $D_4$ ensemble}} ($N{=}4$)$^{\S}$   & test & 39.2 & 59.0 & 35.2 & 22.3 \\
        \textcolor{gray}{SCPNet~\cite{xia2023scpnet} at $\#$frame${=}4^{\ddagger}$} & \textcolor{gray}{test} & \textcolor{gray}{47.5} & \textcolor{gray}{68.5} & \textcolor{gray}{43.9} & \textcolor{gray}{31.0} \\
        \midrule
        PaSCo~\cite{cao2024pasco}$^{\dagger}$ ($3$ subnets) & val  & 30.1 & \na  & \na & \na \\
        $\bullet$\,SGSC (from noise, $T{=}100$)                & val  & 30.5 & 53.6 & 25.6 & 8.5 \\
        $\bullet$\,SGSC\,$+$\,S\textsuperscript{2}D\textsuperscript{2} & val & 32.8 & 55.4 & 28.3 & 12.5 \\
        SCPNet~\cite{xia2023scpnet} (released weights)$^{\sharp}$ & val & 37.55 & 50.24 & \na & \na \\
        \;\;$+$\,SemCity~\cite{lee2024semcity}$^{\sharp}$ & val & 38.19 & \textbf{59.25} & \na & \na \\
        SCPNet$^{*}$ (our port, frozen)             & val  & 36.17 & 49.9 & 32.5 & 14.7 \\
        $\bullet$\,SCPNet$^{*}$\,$+$\,S\textsuperscript{2}D\textsuperscript{2} ($N{=}1$) & val & \textbf{38.54} & 52.7 & \textbf{35.2} & \textbf{19.6} \\
        \midrule
        \multicolumn{6}{@{}p{\columnwidth}@{}}{\scriptsize SC-m: safety-critical mIoU. \textbf{Bold} = best per column within a split; $\bullet$~ours. $^{*}$our port; $^{\P}$point-cloud output, voxelised; $^{\sharp}$quoted from~\cite{lee2024semcity} Tab.~2 on SCPNet's \emph{released} weights, its base sits $1.4$\,pp above our port. Excluded from bolding: $^{\S}$eight-view $D_4$ ensemble, $^{\dagger}$panoptic-mode evaluator, $^{\ddagger}$four sweeps, $^{\parallel}$test-time adaptation.} \\
        \bottomrule
    \end{tabular}
\end{table}

\Cref{tab:main_results} places our rows against published SSC results on both splits. On the hidden test leaderboard, one $N{=}1$ correction step over a frozen SCPNet base reaches $38.8\%$ mIoU: $+2.1$\,pp over the published score of the base it refines. To our knowledge that is the strongest such test-leaderboard result to date. On val the same checkpoint reaches $38.54\%$ against its base's $36.17\%$. The nearest comparison is SemCity~\cite{lee2024semcity}, which refines the same base with a continuous prior: it reports $38.19\%$ from SCPNet's released weights, which score $1.4$\,pp above our port, and it gains far more completion IoU than we do ($+9.0$ against $+2.8$). Our lead is $0.35$\,pp on mIoU from a weaker anchor, so we read the two as comparable rather than ranked. Run from noise instead of from a base, the same generative machinery (SGSC, $\mathbf{c}=\{\mathbf{B},\mathbf{F}\}$, full posterior sampling) reaches $30.5\%$ with no base at all, and $32.8\%$ once its own output is corrected. We set that beside PaSCo~\cite{cao2024pasco} without differencing it: its evaluator is not protocol-matched to ours, and our predicate admits its single subnet ($28.2\%$) rather than the three-subnet $30.1\%$. Among prior generative SSC, only DiffSSC~\cite{cao2024diffssc} has an official-protocol SemanticKITTI entry, $27.4\%$ on the hidden test; Lee~\textit{et al.}~\cite{lee2023diffusion}, whose from-noise setting SGSC adopts, report only on synthetic CarlaSC, and OccGen~\cite{wang2024occgen}'s SemanticKITTI entry is monocular-camera SSC with the LiDAR stream removed. SGSC's $32.8\%$ is $3.3$\,pp short of the discriminative base, and its conditioning still reads a frozen segmentation backbone.

That margin is indexed on the \emph{causal, single-sweep, single-sample} predicate of \S\,\ref{sec:intro}, and is measured against a base we did not re-submit (test-margin caveats in Appendix~G).

\begin{table}[!t]
    \centering
    \footnotesize
    \caption{\textbf{Per-class val IoU against the frozen base} (seq.~$08$, $N{=}1$). ``Released'': shipped checkpoint ($38.54\%$); ``Retrain'': from-scratch ($38.05\%$).}
    \label{tab:perclass_delta}
    \begin{tabular}{@{}lrrr@{}}
        \toprule
        Class            & Base IoU & Released $\Delta$ & Retrain $\Delta$ \\
        \midrule
        car              & $50.3$ & $+1.1$ & $+1.0$ \\
        bicycle          & $23.7$ & $+0.6$ & $-0.1$ \\
        motorcycle       & $35.6$ & $-0.1$ & $+0.2$ \\
        truck            & $54.7$ & $+5.3$ & $+5.4$ \\
        other-vehicle    & $42.0$ & $+2.5$ & $+2.3$ \\
        person           & $22.0$ & $+1.2$ & $+1.9$ \\
        bicyclist        & $18.0$ & $+5.3$ & $+5.2$ \\
        \textbf{motorcyclist} & $\mathbf{4.1}$ & $\mathbf{+8.3}$ & $\mathbf{+0.3}$ \\
        road             & $70.0$ & $+4.6$ & $+4.3$ \\
        parking          & $59.3$ & $+2.4$ & $+2.3$ \\
        sidewalk         & $51.9$ & $+2.1$ & $+1.9$ \\
        other-ground     & $13.6$ & $+0.4$ & $+0.1$ \\
        building         & $33.3$ & $+1.5$ & $+0.7$ \\
        fence            & $29.9$ & $+0.3$ & $+0.1$ \\
        vegetation       & $38.5$ & $+2.2$ & $+1.9$ \\
        trunk            & $32.0$ & $+0.4$ & $+0.2$ \\
        terrain          & $51.9$ & $+2.6$ & $+2.0$ \\
        pole             & $37.7$ & $+0.1$ & $+0.1$ \\
        traffic-sign     & $18.8$ & $+4.2$ & $+5.8$ \\
        \midrule
        \textbf{Overall mIoU} & $\mathbf{36.17}$ & $\mathbf{+2.36}$ & $\mathbf{+1.9}$ \\
        VRU-IoU          & $14.7$ & $+4.9$ & $+2.5$ \\
        \bottomrule
    \end{tabular}
\end{table}

\Cref{tab:perclass_delta} breaks that val gain down by class, and it is not spread evenly. It concentrates where the frozen base is weakest and the missing geometry is still recoverable from context: bicyclist and truck ($+5.3$ each), road ($+4.6$), traffic-sign ($+4.2$), terrain ($+2.6$), other-vehicle ($+2.5$) and parking ($+2.4$). It is flat on thin structures that give a correction step almost nothing to close, at pole ($+0.1$), fence ($+0.3$), trunk and other-ground ($+0.4$); and on the static two-wheelers, bicycle ($+0.6$) and motorcycle ($-0.1$). Eighteen of nineteen classes improve. The exception, motorcycle, loses a tenth of a point. So on this base the operator lifts the tail without trading away what it already handles, and it does so class-selectively rather than by uniformly sharpening the grid. That pattern is specific to SCPNet; \cref{sec:portable_s2d2} reports where it does not hold.

\looseness=-1 One entry in that column we do not claim. Motorcyclist's $+8.3$\,pp fails to reproduce twice: a from-scratch retrain recovers $+0.3$, and on the hidden test the class rises only $1.8\!\to\!3.1$ against SCPNet's published row. We therefore read the validation figure as a seed-and-split artefact and print the retrain column beside it, where the other eighteen classes agree within $1.6$\,pp. The safety metrics inherit the same discount: VRU-IoU's $+4.9$ becomes $+2.5$ on retrain, and the safety claim should be read against that.

\begin{figure*}[!tbp]
    \centering
    \includegraphics[width=0.92\textwidth]{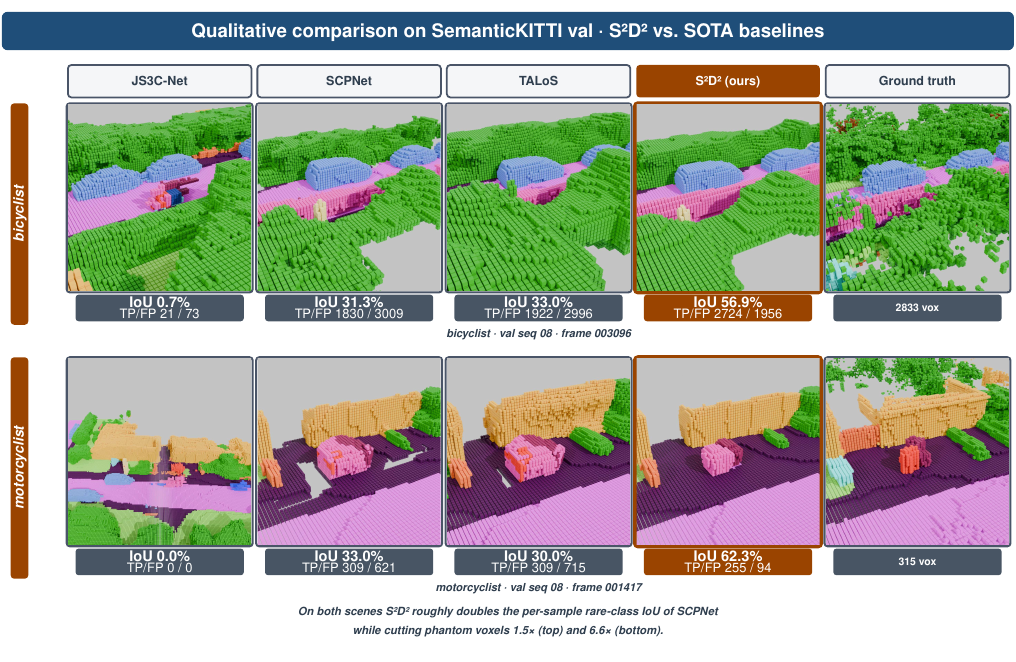}
    \caption{\textbf{Qualitative comparison on val seq.~08.} Three frozen sources (JS3C-Net, SCPNet, TALoS's adaptation of SCPNet), our S\textsuperscript{2}D\textsuperscript{2} on the SCPNet base, and ground truth; \textbf{top} bicyclist, \textbf{bottom} motorcyclist. Chips give per-sample rare-class IoU and TP/FP voxel counts (from the $N{=}4$, $+D_4$-TTA configuration).}
    \label{fig:qualitative}
\end{figure*}

\begin{figure*}[!tbp]
    \centering
    \setlength{\tabcolsep}{1.5pt}
    \renewcommand{\arraystretch}{0.6}
    \newcommand{\qrow}[1]{%
        \includegraphics[width=0.21\textwidth]{material/rendered/p1q1/#1_input_p.png} &
        \includegraphics[height=50.5pt]{material/rendered/p1q1/#1_bev_p.png} &
        \includegraphics[width=0.21\textwidth]{material/rendered/p1q1/#1_gssc_p.png} &
        \includegraphics[width=0.21\textwidth]{material/rendered/p1q1/#1_gt_p.png}}
    \begin{tabular}{@{}cccc@{}}
        \qrow{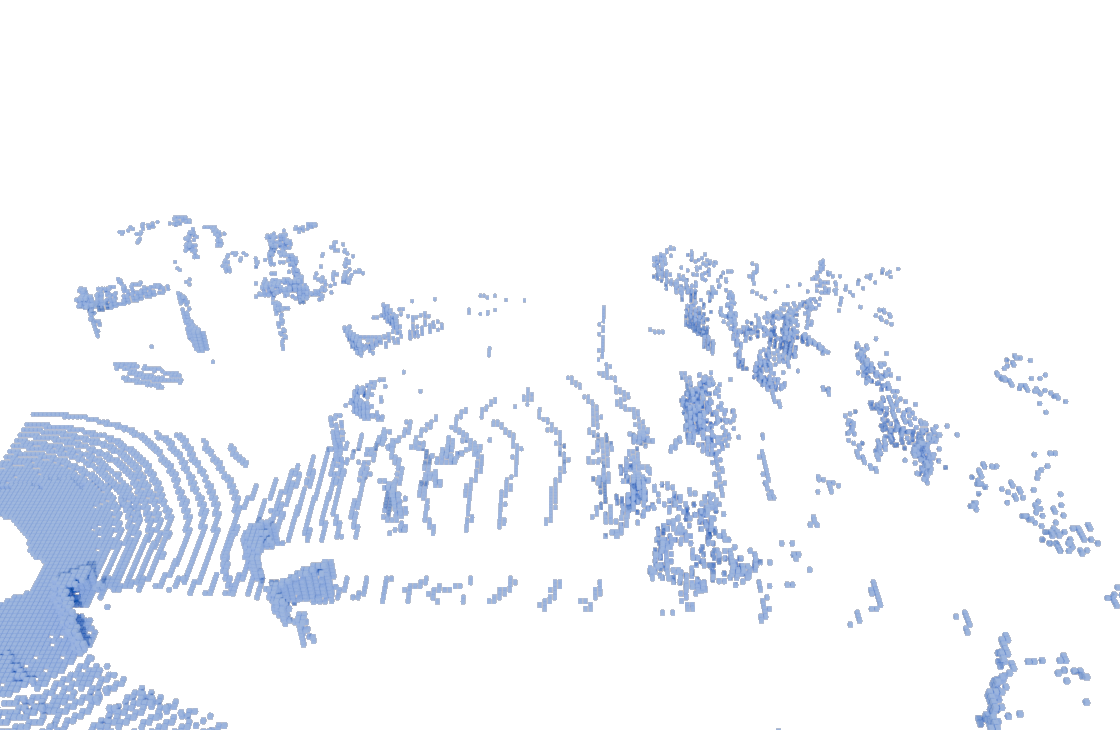} \\
        \qrow{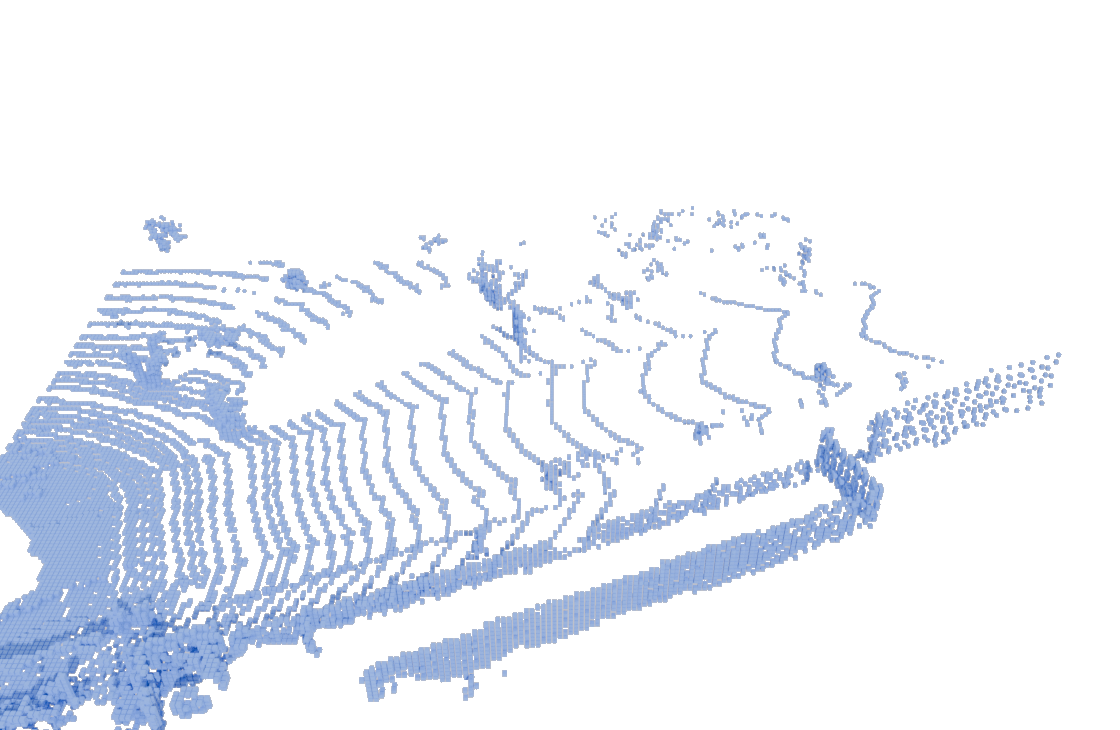} \\[2pt]
        \scriptsize (a) sparse LiDAR input &
        \scriptsize (b) BEV stream $\mathbf{B}$ &
        \scriptsize (c) our completion &
        \scriptsize (d) ground truth \\
    \end{tabular}
    \caption{\textbf{Our completions on val seq.~08} (rows: frames 004025, 001850). From a ${\sim}1\%$-occupied sweep (a) and the base-derived BEV (b), one correction step recovers dense semantics (c) close to ground truth (d).}
    \label{fig:completions}
\end{figure*}

\looseness=-1 \Cref{fig:qualitative} shows what the correction does to a scene. All three frozen sources already recover the road surface and the building façades, and one correction step leaves that bulk alone: road, sidewalk, building and vegetation drift by under $1\%$ IoU. The difference is at the tail. Where the base returns a fragmentary, under-filled rider, one step closes it into a coherent object with sharper boundaries, roughly doubling per-sample rare-class IoU with ${\approx}6.6{\times}$ fewer phantom voxels on the motorcyclist scene, while JS3C-Net collapses on both. The per-voxel figures in this paragraph are read off chips rendered from the $N{=}4$, $+D_4$-TTA configuration, so they characterise the correction rather than the $N{=}1$ operating point the headline reports; this is the behaviour \cref{eq:correction_step} predicts. \Cref{fig:completions} shows the same single step on two ordinary frames, from a ${\sim}1\%$-occupied sweep to dense semantics. \Cref{fig:qualitative_gallery} extends this to six rare-class scenes; the per-sample voxel counts behind the chips are in Appendix~D.

\subsection{\texorpdfstring{S\textsuperscript{2}D\textsuperscript{2}}{S2D2} as a Base-Agnostic Refinement Operator}
\label{sec:portable_s2d2}

One operator, three frozen bases, one pass each. S\textsuperscript{2}D\textsuperscript{2} is trained once per base on that base's own decoded output and then run for a single $N{=}1$ correction step. No base is retrained or architecturally touched. It lifts every one: LMSCNet by $+1.8$, JS3C-Net by $+1.6$, SCPNet by $+2.36$\,pp val mIoU (\cref{tab:portable_s2d2}). That is what \emph{base-agnostic} means. The three span the field: a lightweight 2D--3D hybrid, a point-voxel segmentation-coupled network, and a distillation-trained state of the art. The denoiser and recipe are fixed across them, each base refined under its own best pool, step budget and LiDAR stream, so the controlled quantity is the per-base uplift rather than a cross-base absolute.

\begin{table}[!t]
    \centering
    \scriptsize
    \setlength{\tabcolsep}{3pt}
    \caption{\textbf{One denoiser architecture on three frozen bases}, each trained on its own base's output and refined by a single $N{=}1$ pass. Val seq.~$08$; $\Delta$ is the mIoU uplift over the paired base.}
    \label{tab:portable_s2d2}%
    \label{tab:main_results_t2}
    \begin{tabular}{@{}lccccc@{}}
        \toprule
        Configuration & mIoU & Comp & $\Delta$ & SC-m & VRU \\
        \midrule
        LMSCNet$^{\flat}$                                       & 14.8 & 46.7 & --     & 10.4 & 0.0 \\
        \;$+$S\textsuperscript{2}D\textsuperscript{2}          & \textbf{16.6} & 46.7 & \textbf{+1.8} & \textbf{12.1} & \textbf{0.3} \\
        JS3C-Net$^{*\flat}$                                    & 22.7 & \textbf{53.1} & --     & 18.6 & \textbf{3.8} \\
        \;$+$S\textsuperscript{2}D\textsuperscript{2}$^{\dagger}$ & \textbf{24.3} & 50.9 & \textbf{+1.6} & \textbf{20.0} & 2.4 \\
        SCPNet$^{*}$                                           & 36.17 & 49.9 & --     & 32.5 & 14.7 \\
        \;$+$S\textsuperscript{2}D\textsuperscript{2}          & \textbf{38.54} & \textbf{52.7} & \textbf{+2.36} & \textbf{35.2} & \textbf{19.6} \\
        \midrule
        \multicolumn{6}{@{}p{\columnwidth}@{}}{\scriptsize \textbf{Bold} = better within each base pair; the SCPNet pair repeats \cref{tab:main_results} as this table's reference. $^{*}$re-implemented, $^{\flat}$re-scored, $^{\dagger}$official evaluator.} \\
        \bottomrule
    \end{tabular}
\end{table}

Two bases carry caveats worth stating before the numbers are used. The official SCPNet release depends on \texttt{spconv~1.0}, removed from PyPI. We run it under \texttt{spconv~2.3}~\cite{spconv2022} after kernel-shape patches replicating v1's shared-indice-key behaviour, loading \texttt{pretrained.pth} as-is to reach $36.17\%$ val mIoU, $1.0$\,pp under the paper's $37.2\%$.\footnote{\label{fn:scpnet_repro}%
A recent independent study reports both of our reference points under caveats (SCPNet as non-reproducible, TALoS as multi-frame test-time optimisation) and calls its own $28.8\%$ recipe the state of the art among fully reproducible single-frame methods~\cite{martyniuk2026easyboosts}. We report against the published numbers because those are what a reader can check.}
\looseness=-1 We did not re-submit the port. The test row of \cref{tab:main_results} therefore quotes SCPNet's published $36.7\%$, and val $\Delta$ is measured against $36.17\%$ throughout (port details in Appendix~G).

JS3C-Net marks the boundary of the transfer. Under the protocol-matched official evaluator refinement takes it from $22.7$ to $24.3\%$, but not uniformly. The gain holds on ten of nineteen classes while two of the three VRU classes \emph{lose} ground: person $8.7\!\to\!5.9$ and bicyclist $2.6\!\to\!1.2$, motorcyclist $0.0$ throughout, so VRU-IoU falls $3.8\!\to\!2.4$. It buys bulk geometry at the tail's expense. The synthetic pool that helps voxel-grid bases also costs this one $-3.8$\,pp under our internal evaluator, its point-voxel head treating the voxel-derived cloud as out of distribution. On both axes the operator's benefit is scoped to voxel-grid-native bases rather than to LiDAR architectures in general. A second refinement round does not repair this or anything else: retrained from scratch on the SCPNet base's own Round-1 prediction ($38.73\%$ val at $N{=}4{+}D_4$ TTA), over the real sequences alone, it reaches $37.7\%$ val mIoU, $1.1$\,pp \emph{below} it, over-smoothing an already-refined source.

\looseness=-1 The precision/recall character of the correction is visible across bases. On the over-firing LMSCNet base the semantic gain comes from \emph{pruning} false-positive occupancy, so completion IoU is unchanged at $46.7$. On JS3C-Net the same trade costs $2.2$\,pp of completion IoU where it does not cancel. On SCPNet it lifts both ($49.9\!\to\!52.7$). Two per-class regressions replicate across the weaker bases and are worth naming: building falls $34.7\!\to\!28.6$ on LMSCNet and $36.6\!\to\!30.6$ on JS3C-Net, the largest single-class loss on each, and LMSCNet also loses truck ($1.9\!\to\!0.4$). On a weaker base the correction does not simply add. On the bulk structural classes it can subtract. An operating-point reading is not excluded by any of this: the base recovers only $53.1\%$ of the occupied voxels, so a confidence-ranked densification of it could shift the same trade. Ruling that out needs the base's per-voxel posteriors, which its released \texttt{argmax} predictions do not carry (Appendix~E). Occupancy forensics and the full per-class breakdown for all three bases are in Appendix~D.

\subsection{Ablations}
\label{sec:ablations}
\label{sec:additional_ablations}
\label{sec:sgsc_ablation}
\label{sec:loss_ablation}
\label{sec:refiner_control}
\label{sec:bev_ablation}
\label{sec:single_frame_retrain_ablation}
\label{sec:fromnoise_ablation}
\label{sec:step_reduction}
\label{sec:ema_ablation}
\label{sec:computation}

\begin{table}[!t]
    \centering
    \scriptsize
    \setlength{\tabcolsep}{3.5pt}
    \caption{\textbf{Ablations.} Most rows remove or replace a single element of the headline recipe; where a row differs in more than one, the text says so. Each block names its own reference and probe.}
    \label{tab:ablations}
    \begin{tabular}{@{}lcc@{}}
        \toprule
        Configuration & mIoU & $\Delta$ \\
        \midrule
        \multicolumn{3}{@{}l}{\emph{SGSC conditioning} (from noise, oracle GT BEV, $100$ frames):} \\
        no conditioning                                  & 1.29  & $-41.0$ \\
        BEV stream $\mathbf{B}$ only                     & 34.08 & $-8.2$ \\
        $\mathbf{B}+$LSK3D features $\mathbf{F}$ (ref.)  & \textbf{42.30} & -- \\
        \midrule
        \multicolumn{3}{@{}l}{\emph{S\textsuperscript{2}D\textsuperscript{2} components} ($100$-frame probe, $\Delta$ vs.\ frozen base):} \\
        no BEV stream                                    & \na & $+1.1$ \\
        no synthetic pool                                & \na & ${\sim}+1.7$ \\
        probe reference (BEV $+$ $32$K pool)             & \na & $\mathbf{+2.5}$ \\
        $+$ 3D multi-scale stream                        & \na & $+2.5$ \\
        \midrule
        \multicolumn{3}{@{}l}{\emph{S\textsuperscript{2}D\textsuperscript{2} loss} (full val, $\Delta$ vs.\ headline):} \\
        CE on $\hat{\mathbf{x}}_0$, no KL term            & 10.3  & $-28.2$ \\
        KL term alone                                     & 36.1  & $-2.5$ \\
        KL $+$ Lov\'asz $+$ focal (ref.)                  & \textbf{38.54} & -- \\
        \midrule
        \multicolumn{3}{@{}l}{\emph{Training path} (full val; reference per row, see notes):} \\
        fitted at the endpoint (supervised control)       & 27.0  & $-11.5$ \\
        trained only at $t{=}T$                           & 32.74 & $-5.4^{\natural}$ \\
        \midrule
        \multicolumn{3}{@{}l}{\emph{Deployment} (full val, $\Delta$ vs.\ headline $N{=}1$):} \\
        $N{=}1$, no TTA (ref.)                            & \textbf{38.54} & -- \\
        $N{=}4$ (interior peak)                           & 38.65 & $+0.11$ \\
        $N{=}100$ (full schedule)                         & 38.2  & $-0.34$ \\
        without EMA weights$^{\sharp}$                          & \na   & $-3.5$ \\
        \midrule
        \multicolumn{3}{@{}p{\columnwidth}@{}}{\scriptsize $^{\natural}$released codebase at $40$K, against the matched full-schedule arm's $38.1\%$. $^{\sharp}$$100$-frame probe at step~$35$K, $N{=}100$, not full val. \na: the probe stores aggregate deltas only. ${\sim}$: probe precision; Appendix~C gives the pool as a $+0.8$ marginal.} \\
        \bottomrule
    \end{tabular}
\end{table}

Most rows of \cref{tab:ablations} change one element of the recipe and rescore; each block states its own reference, because the blocks use different probes (provenance and full sweeps in Appendix~C).

\begin{figure*}[!t]
    \centering
    \setlength{\tabcolsep}{1.5pt}
    \begin{tabular}{@{}ccc@{}}
        \includegraphics[width=0.325\textwidth]{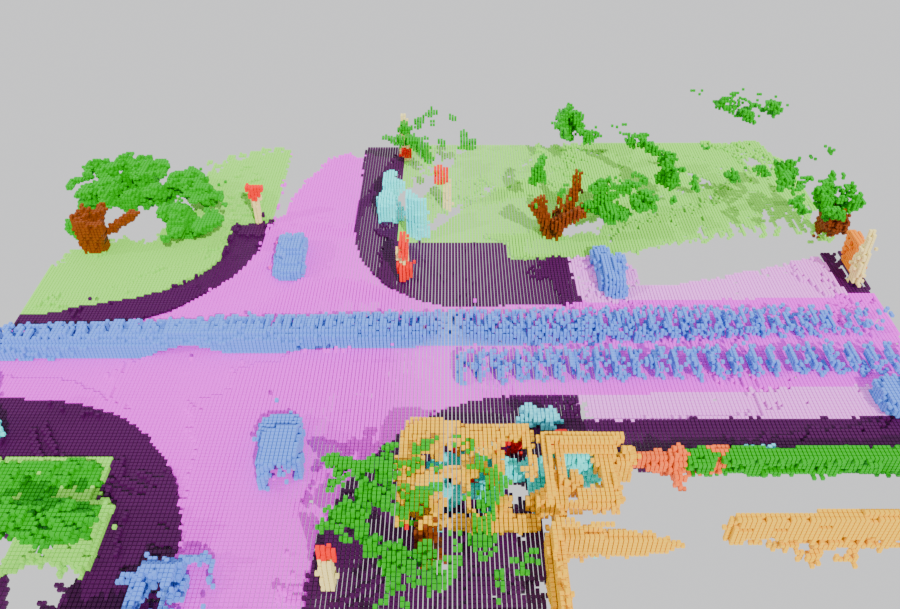} &
        \includegraphics[width=0.325\textwidth]{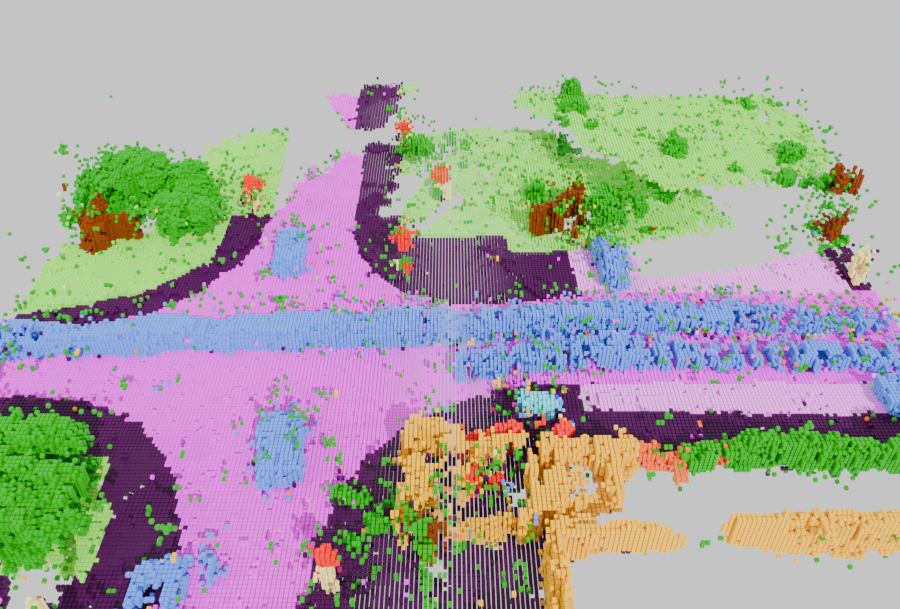} &
        \includegraphics[width=0.325\textwidth]{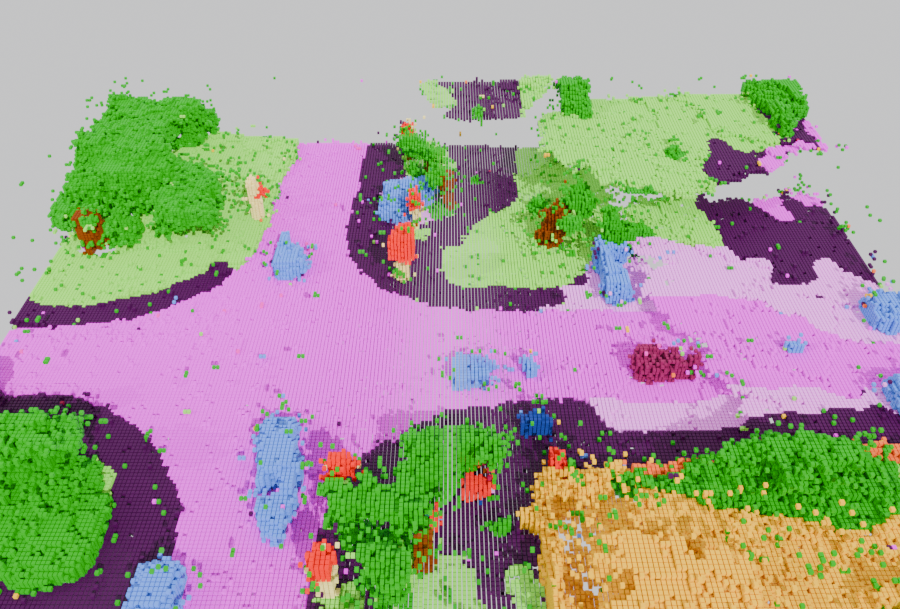} \\[-1pt]
        {\scriptsize Ground-truth scene} &
        {\scriptsize SGSC completion, \emph{oracle} BEV} &
        {\scriptsize SGSC completion, \emph{predicted} BEV} \\
    \end{tabular}
    \caption{\textbf{The generator inherits the quality of its guidance.} Ground-truth scene (left) and two $256^2{\times}32$ SGSC completions (val seq.~08, frame~$6$) from one development checkpoint (from-noise $T{=}100$, identical weights, only the BEV \emph{conditioning} differs: oracle vs.\ predicted). A predicted BEV keeps the road surface but collapses the road users.}
    \label{fig:sgsc_conditioning}
\end{figure*}

\emph{Conditioning carries the generator.} Removing it entirely leaves SGSC barely off the floor at $1.29\%$; the BEV stream alone carries it to $34.08\%$, and the per-voxel feature stream adds a further $+8.2$ to an oracle ceiling of $42.30\%$. These are oracle-BEV ceilings, not deployable scores. The deployed SGSC of \cref{tab:main_results} uses its own BEV head and reaches $30.5\%$. Guidance \emph{quality} matters as much as its presence: swapping the oracle BEV for a predicted one at fixed weights leaves bulk geometry intact while the two-wheeler classes collapse, as \cref{fig:sgsc_conditioning} shows side by side. Neither module manufactures semantics that the conditioning does not carry.

\emph{The BEV stream is the dominant component of the refiner, and the synthetic pool the second.} Removing the base-derived BEV stream drops the uplift from $+2.5$ to $+1.1$\,pp; removing the $32$K synthetic pool drops it to ${\sim}+1.7$. Data- and model-side contributions are both positive and do not cancel. Two negatives sharpen the reading: adding a 3D multi-scale conditioning stream on top is a null result within noise, so we keep the BEV stream alone. Applied to TALoS's already-adapted SCPNet the operator lifts least ($+0.9$), that base having already moved some of the mass the correction targets. Pool size does not scale naively ($32$K is the best measured point), and conditioning on a \emph{refined} rather than base-derived BEV leaves quality inside headline noise, so we use the cheaper unrefined stream.

\emph{The KL posterior term is necessary and not sufficient.} \looseness=-1 Replacing it with cross-entropy on the clean target collapses the model to $10.3\%$, $25.9$\,pp \emph{below} its own input. Yet KL without the class-balanced surrogates improves nothing: it reaches $36.1\%$, the base's $36.17\%$ to within a tenth of a point, stable and inert. The $+2.36$\,pp headline is therefore an interaction, with KL anchoring the per-timestep posterior while those surrogates (Lov\'asz--softmax~\cite{berman2018lovasz} and focal weighting) supply the only gradient that pushes the model off its source. That is why the per-class gain of \cref{tab:perclass_delta} concentrates on the classes those surrogates target.

\looseness=-1 \emph{What the multi-timestep training path buys.} At $N{=}1$ the deployed operator is one forward pass whose source term contributes under $1\%$ of the output ($\bar{\alpha}_T\approx5.6{\times}10^{-3}$), so what does the formulation buy over training the same network to map $\mathbf{x}_{\mathrm{src}}$ to the target in one supervised step? Fitting the endpoint directly is that arm: it floods rather than copies, never reaching the identity map inside its own hypothesis class, so it bounds the question loosely at best. Restricting $t$ to $T$ asks something narrower. We ran it on the released codebase, matched in batch size, seed, data and every loss coefficient. At $40$K steps it reaches $32.74\%$ against a matched full-schedule arm's $38.1\%$, falling $3.4$\,pp below the $36.17\%$ it was handed and flooding the grid rather than copying it. It does not isolate the schedule, though. At $t{=}T$ the state equals $\mathbf{x}_{\mathrm{src}}$ on all but $\bar{\alpha}_T$ of voxels, where both brackets of \cref{eq:s2d2_posterior} collapse onto the source, so the KL yields no gradient; the auxiliary weighting and the $100$ mixtures each pair is seen at move with it. All are consequences of restricting $t$, not choices we could have matched, so the arm prices the variational path \emph{as a whole} and attributes the gap to none of its parts.

\emph{One step suffices at deployment.} Sweeping the correction from $1$ to $100$ steps on full val, quality peaks at a shallow $38.65\%$ at $N{=}4$. Both the $100$-step run and the deployed $N{=}1$ sit within the assumed single-seed band of that peak ($0.45$ and $0.11$\,pp below it). Nothing in the sweep is resolvable, which is the point: one step buys a $100{\times}$ speedup at no measurable cost, and needs neither the retraining nor the distillation that step reduction usually requires~\cite{salimans2022progressive,song2023consistency,zhang2025scorelidar}. EMA weights are required ($-3.5$\,pp without). The headline consumes one sweep at deployment, though its training accumulated several; a retrain matched to single-frame at both reaches $38.4\%$, $0.1$\,pp from the released checkpoint and inside the assumed $0.3$--$0.5$\,pp single-seed band (Appendix~C).

\subsection{Transfer and Secondary Tasks}
\label{sec:transfer}
\label{sec:crossdataset}
\label{sec:synth_quality}
\label{sec:bev_secondary}

\begin{figure*}[!t]
    \centering
    \includegraphics[width=0.62\textwidth]{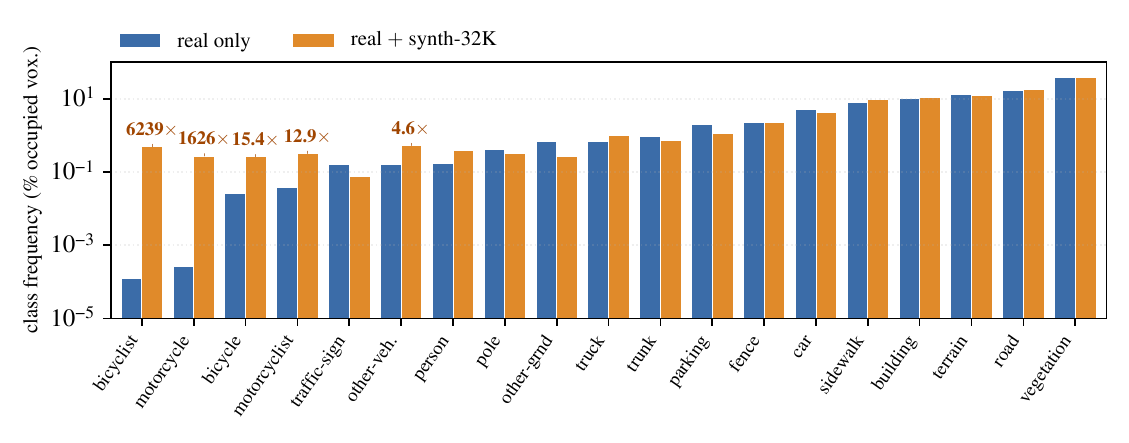}
    \caption{\textbf{PS\textsuperscript{3} rebalances the long tail.} Per-class voxel frequency (log scale, sorted ascending by real-only frequency), real-only vs.\ real$+$synth-32K. The annotated factors are the synthetic pool's own frequency against real-only ($4.6{\times}$--$6{,}239{\times}$); the pooled bars gain $3.2{\times}$--$3{,}907{\times}$.}
    \label{fig:synth_quality}
\end{figure*}

The gains survive outside the training distribution. Refining the frozen SemanticKITTI checkpoint zero-shot, with no target labels and no fine-tuning, lifts SSCBench-KITTI360~\cite{li2024sscbench} from mIoU $5.8$ to $6.2$ and completion IoU $18.1$ to $19.5$, scored over the $16$ shared classes rather than its own $18$. The mIoU delta sits inside the assumed single-seed band, so here the claim rests on completion IoU. On SemanticPOSS~\cite{pan2020semanticposs}, a different sensor, both clear the band: mIoU $1.0\!\to\!6.5\%$ and completion IoU $31.8\!\to\!54.9\%$. Our SemanticPOSS ground truth is reconstructed rather than official, so these are not head-to-head with TALoS. TALoS reports $9.6\%$ from a $7.6\%$ SCPNet base under a SemanticPOSS completion ground truth it does not describe (protocol in Appendix~E).

The synthetic pool that feeds this refiner reshapes the training distribution rather than merely enlarging it: PS\textsuperscript{3} raises rare-class voxel frequency by up to $3{,}907\times$ (\cref{fig:synth_quality}). Its measured benefit is real but bounded, and specific to the voxel-grid-native base we test. Under a retrain matched to single-frame input at both ends, only the pool's peak location transfers, not its margin over real-only (Appendix~C).

\begin{figure*}[!tbp]
    \centering
    \footnotesize
    \setlength{\tabcolsep}{1pt}
    \renewcommand{\arraystretch}{1.05}
    \begin{tabular}{@{}cccc@{}}
        \multicolumn{2}{c}{\textbf{motorcycle} (frame 002614)} & \multicolumn{2}{c}{\textbf{traffic-sign} (frame 002578)} \\[-1pt]
        \multicolumn{2}{c}{\scriptsize{base $4.8\%$ \;\;$\rightarrow$\;\; \textbf{S\textsuperscript{2}D\textsuperscript{2} $\mathbf{44.9\%}$}\quad ($+40.1$)}} & \multicolumn{2}{c}{\scriptsize{base $25.8\%$ \;\;$\rightarrow$\;\; \textbf{S\textsuperscript{2}D\textsuperscript{2} $\mathbf{61.0\%}$}\quad ($+35.2$)}} \\
        \includegraphics[width=0.245\textwidth]{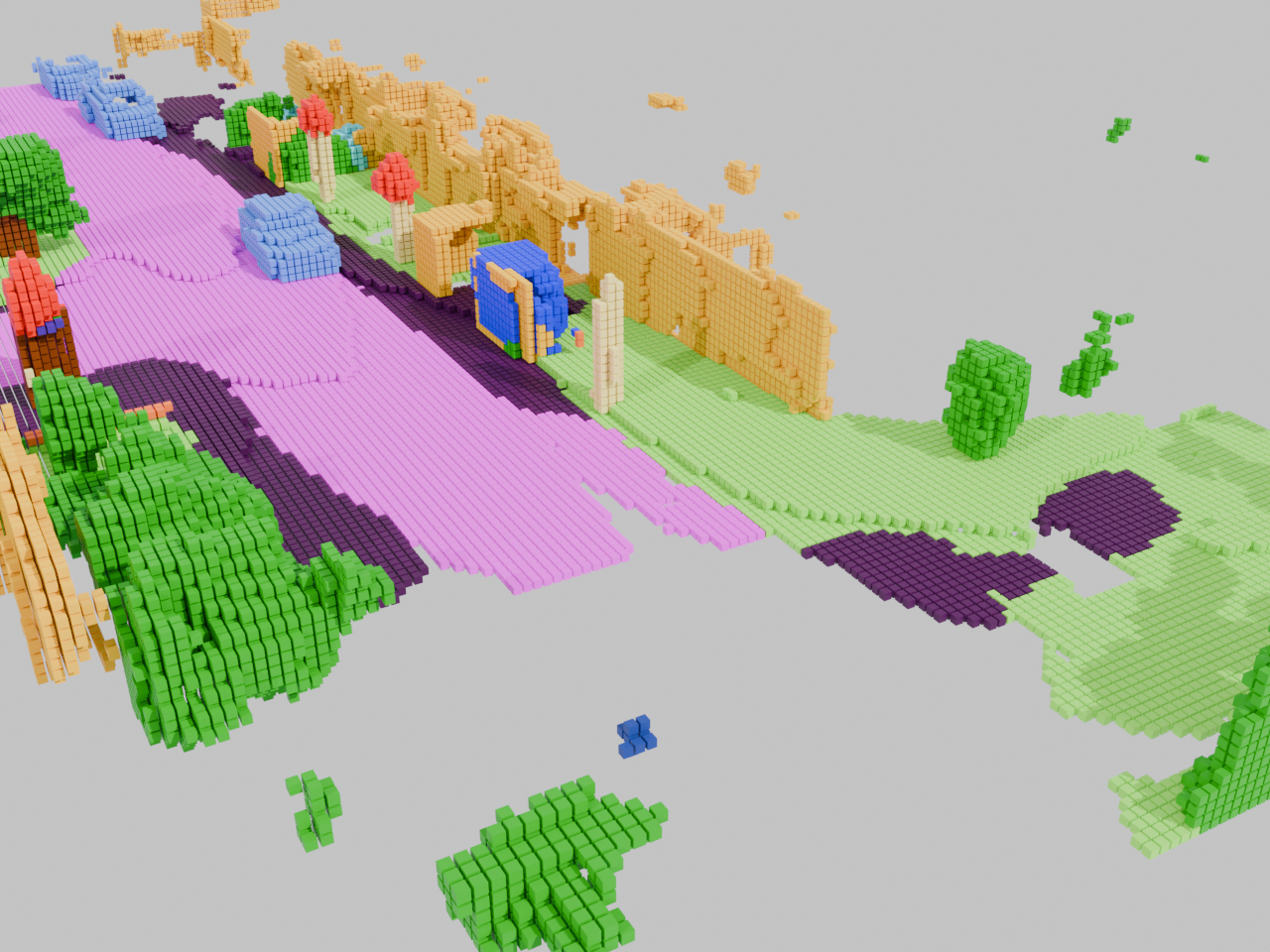} &
        \includegraphics[width=0.245\textwidth]{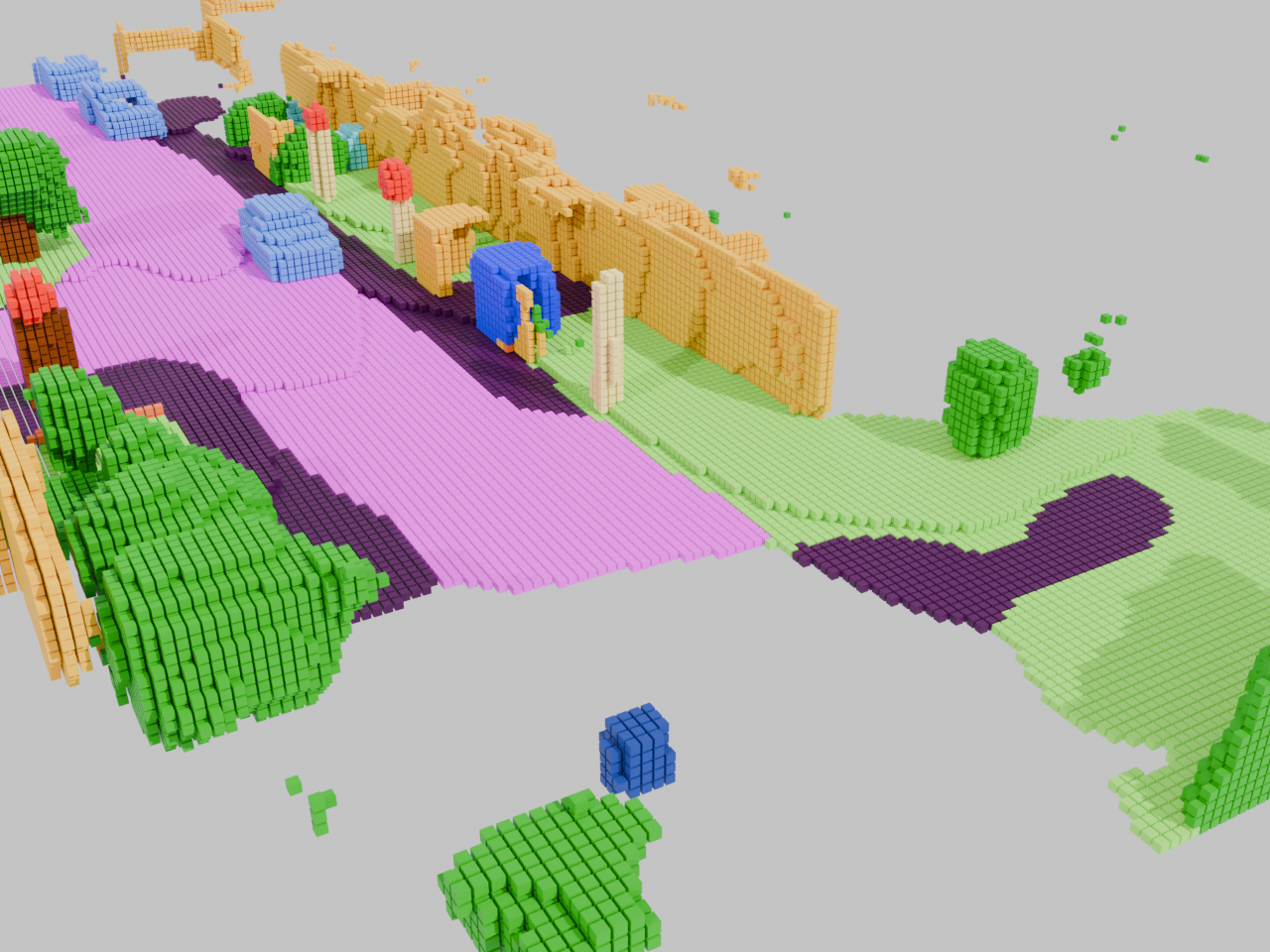} &
        \includegraphics[width=0.245\textwidth]{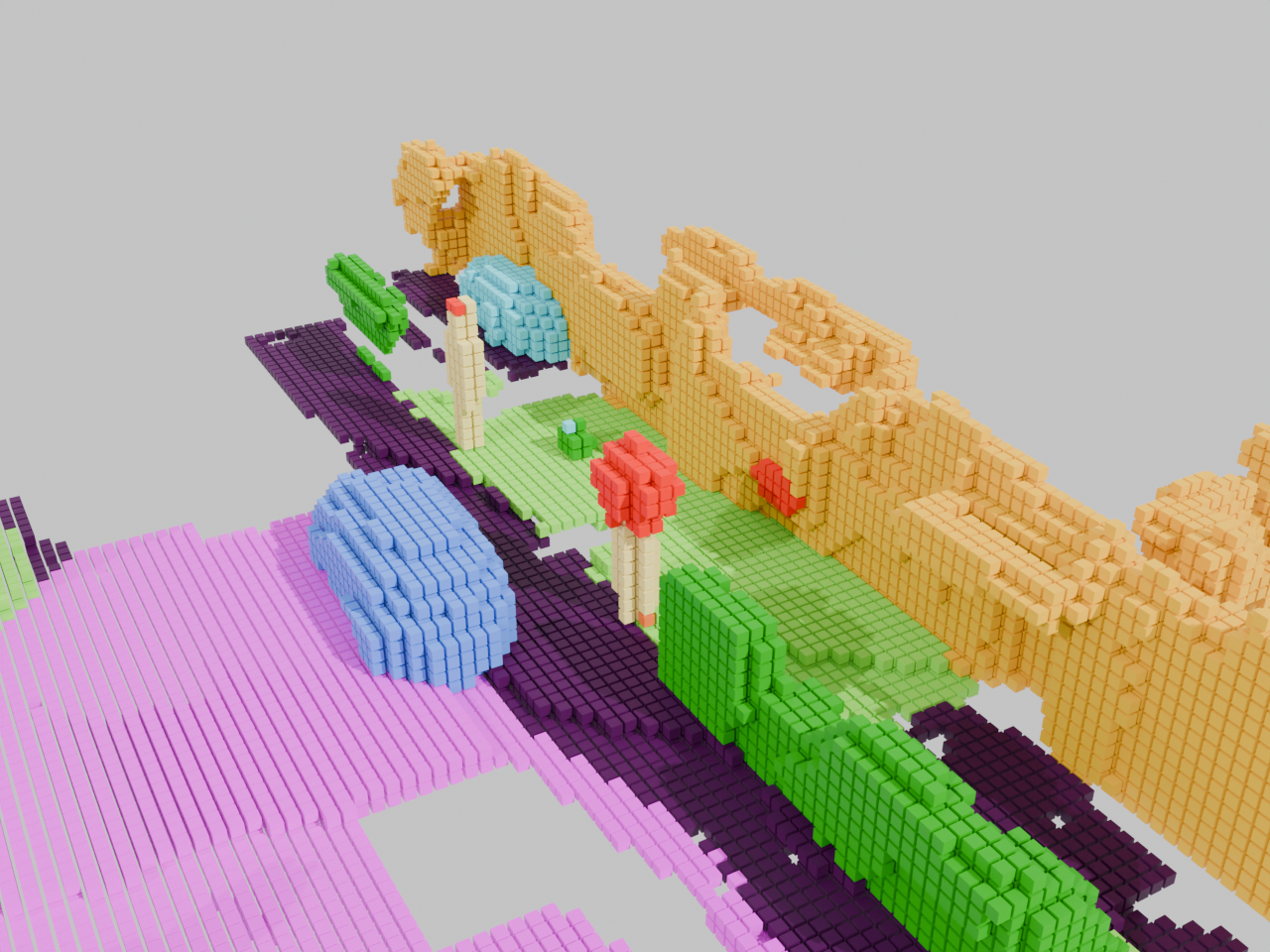} &
        \includegraphics[width=0.245\textwidth]{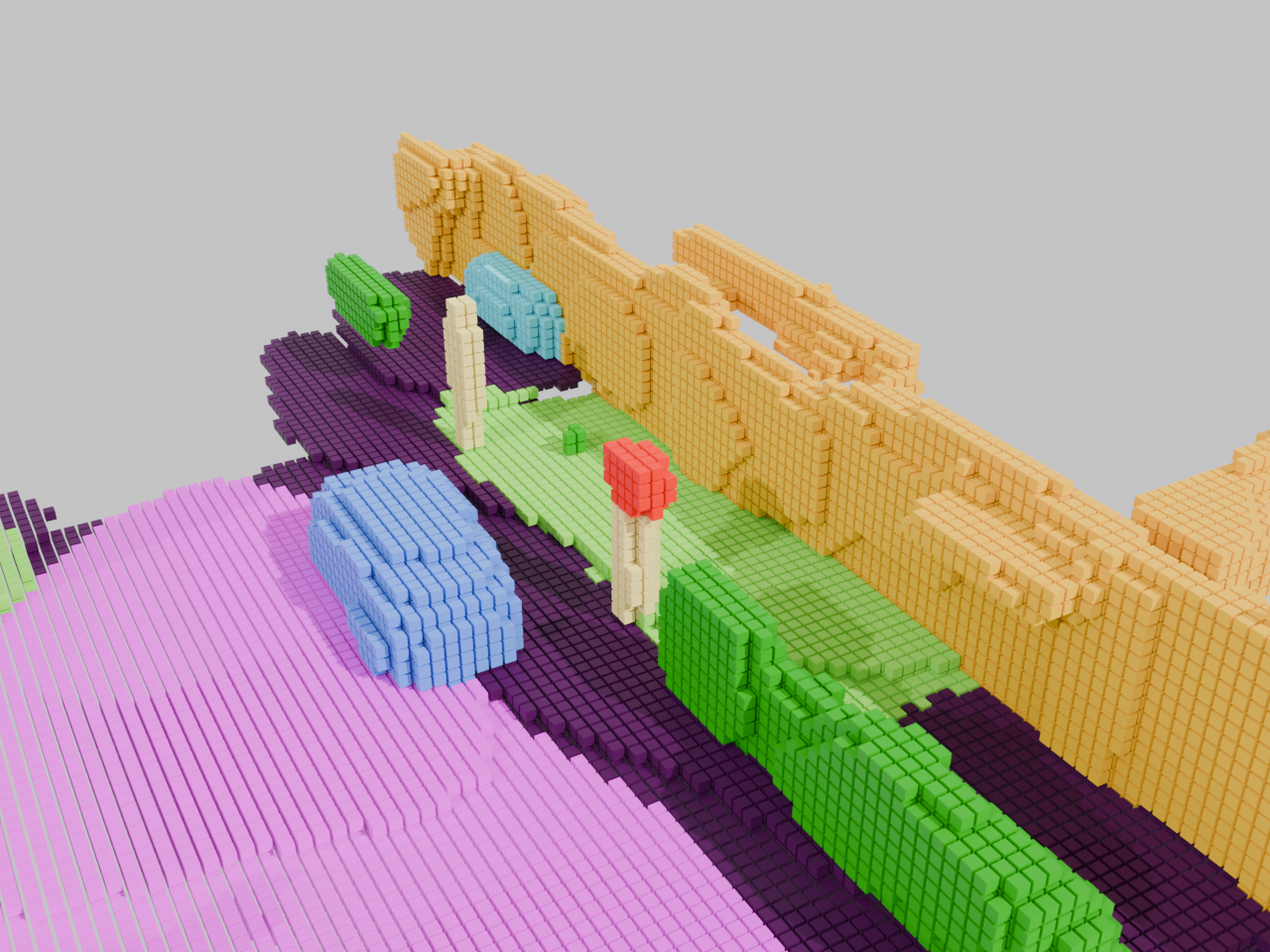} \\
        \multicolumn{2}{c}{\textbf{person} (frame 003364)} & \multicolumn{2}{c}{\textbf{bicyclist} (frame 003103)} \\[-1pt]
        \multicolumn{2}{c}{\scriptsize{base $16.1\%$ \;\;$\rightarrow$\;\; \textbf{S\textsuperscript{2}D\textsuperscript{2} $\mathbf{47.4\%}$}\quad ($+31.3$)}} & \multicolumn{2}{c}{\scriptsize{base $32.0\%$ \;\;$\rightarrow$\;\; \textbf{S\textsuperscript{2}D\textsuperscript{2} $\mathbf{54.6\%}$}\quad ($+22.6$)}} \\
        \includegraphics[width=0.245\textwidth]{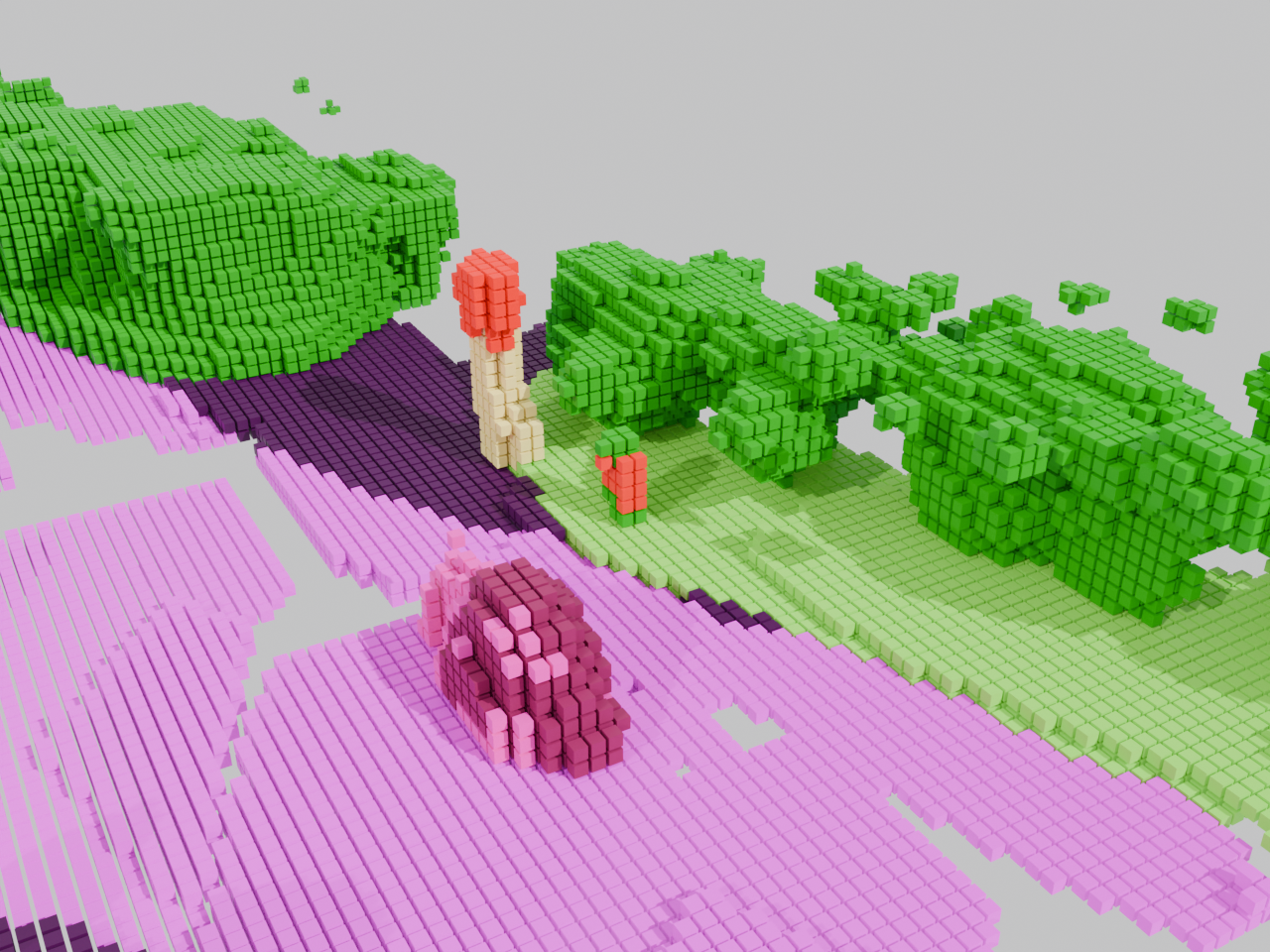} &
        \includegraphics[width=0.245\textwidth]{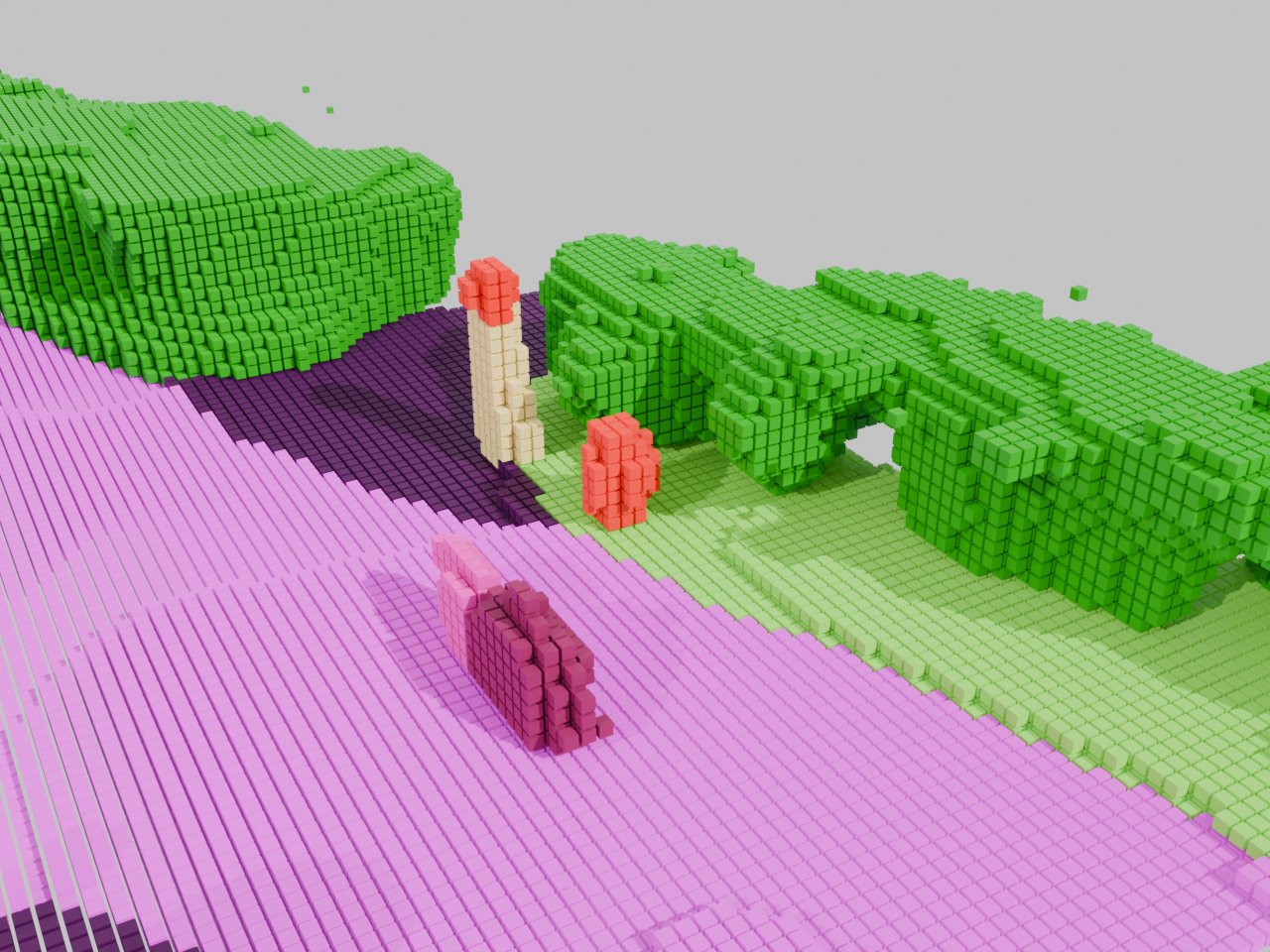} &
        \includegraphics[width=0.245\textwidth]{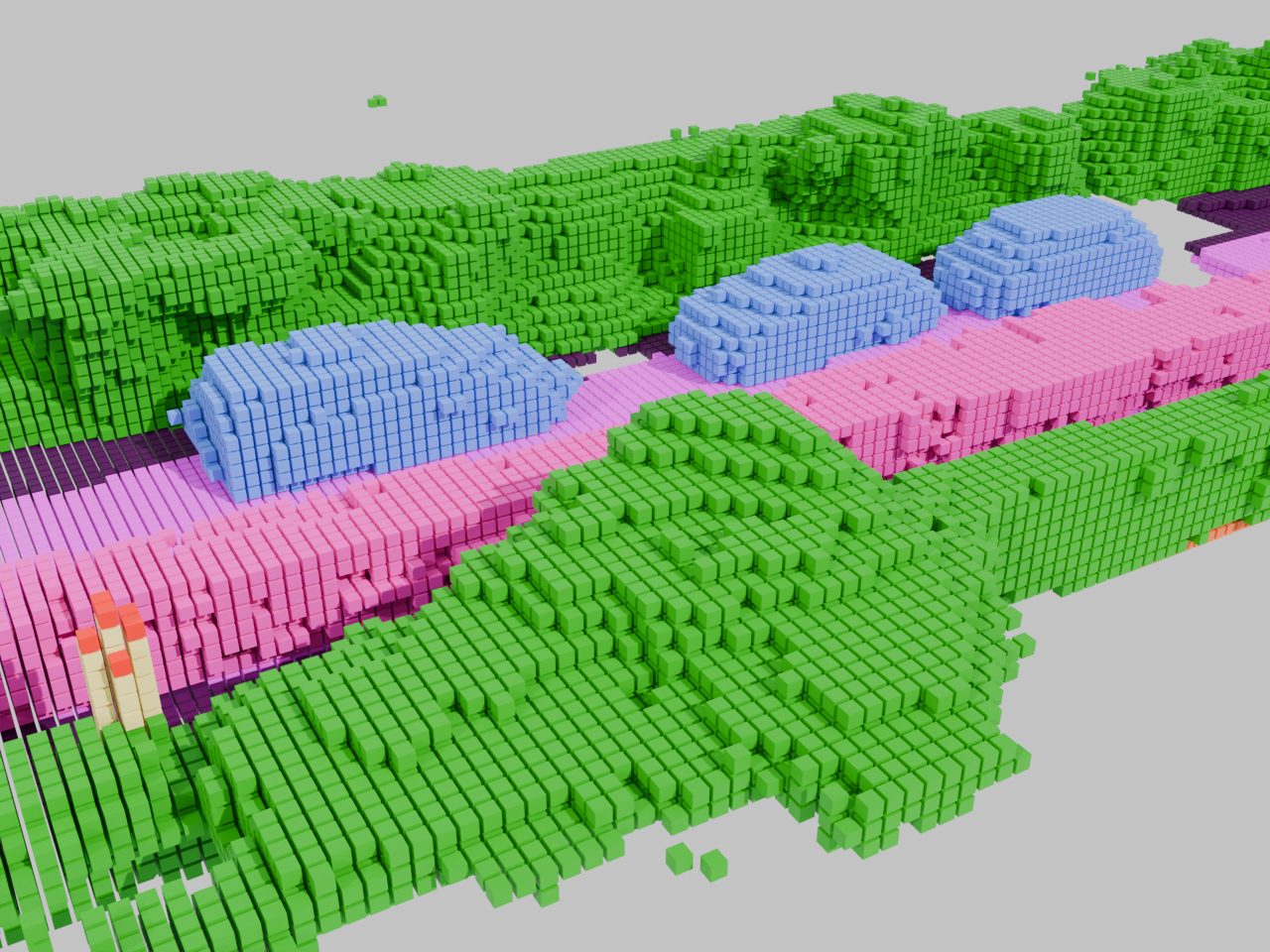} &
        \includegraphics[width=0.245\textwidth]{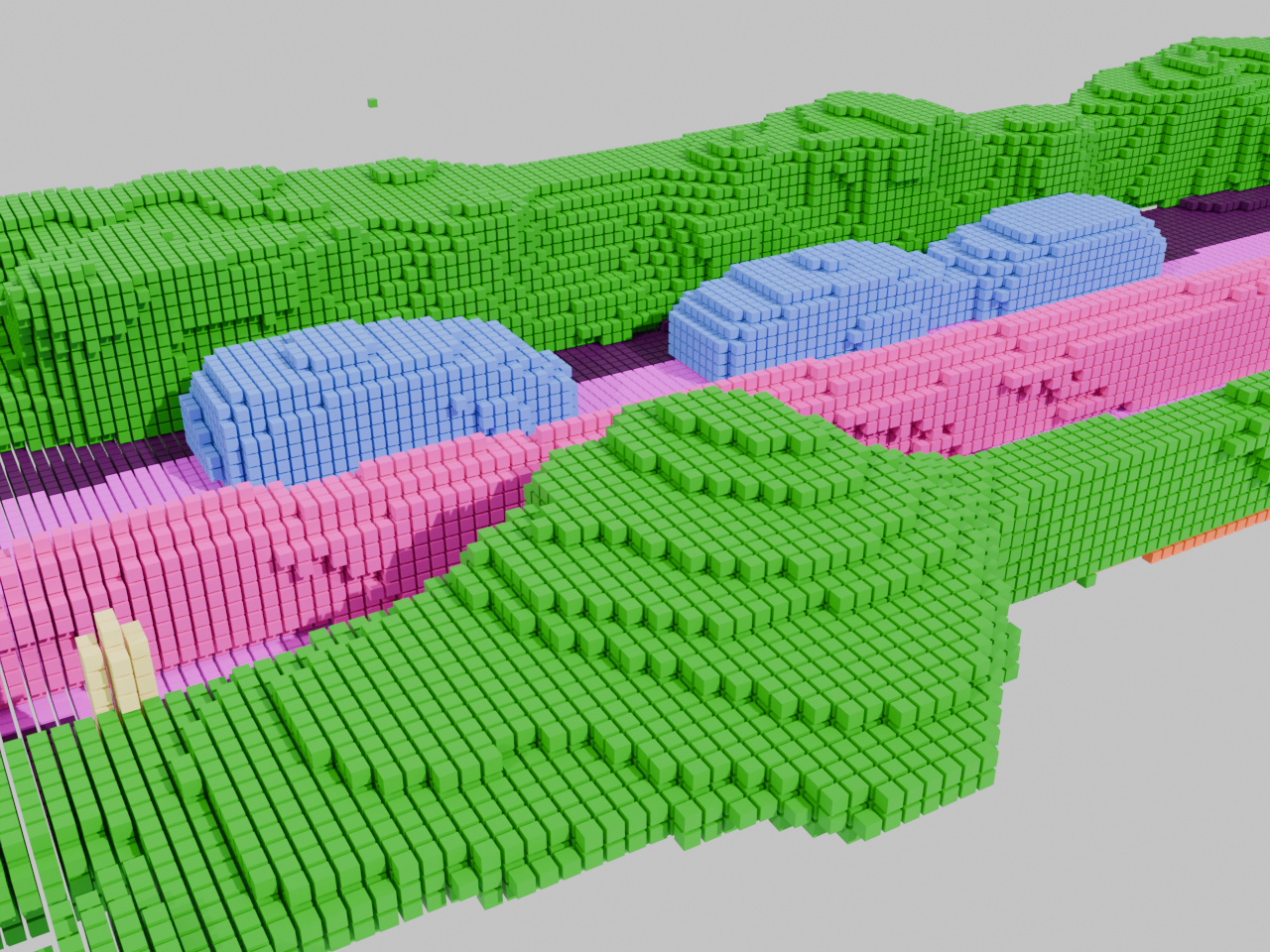} \\
        \multicolumn{2}{c}{\textbf{other-vehicle} (frame 000506)} & \multicolumn{2}{c}{\textbf{bicycle} (frame 001861)} \\[-1pt]
        \multicolumn{2}{c}{\scriptsize{base $52.9\%$ \;\;$\rightarrow$\;\; \textbf{S\textsuperscript{2}D\textsuperscript{2} $\mathbf{74.3\%}$}\quad ($+21.4$)}} & \multicolumn{2}{c}{\scriptsize{base $41.2\%$ \;\;$\rightarrow$\;\; \textbf{S\textsuperscript{2}D\textsuperscript{2} $\mathbf{61.3\%}$}\quad ($+20.1$)}} \\
        \includegraphics[width=0.245\textwidth]{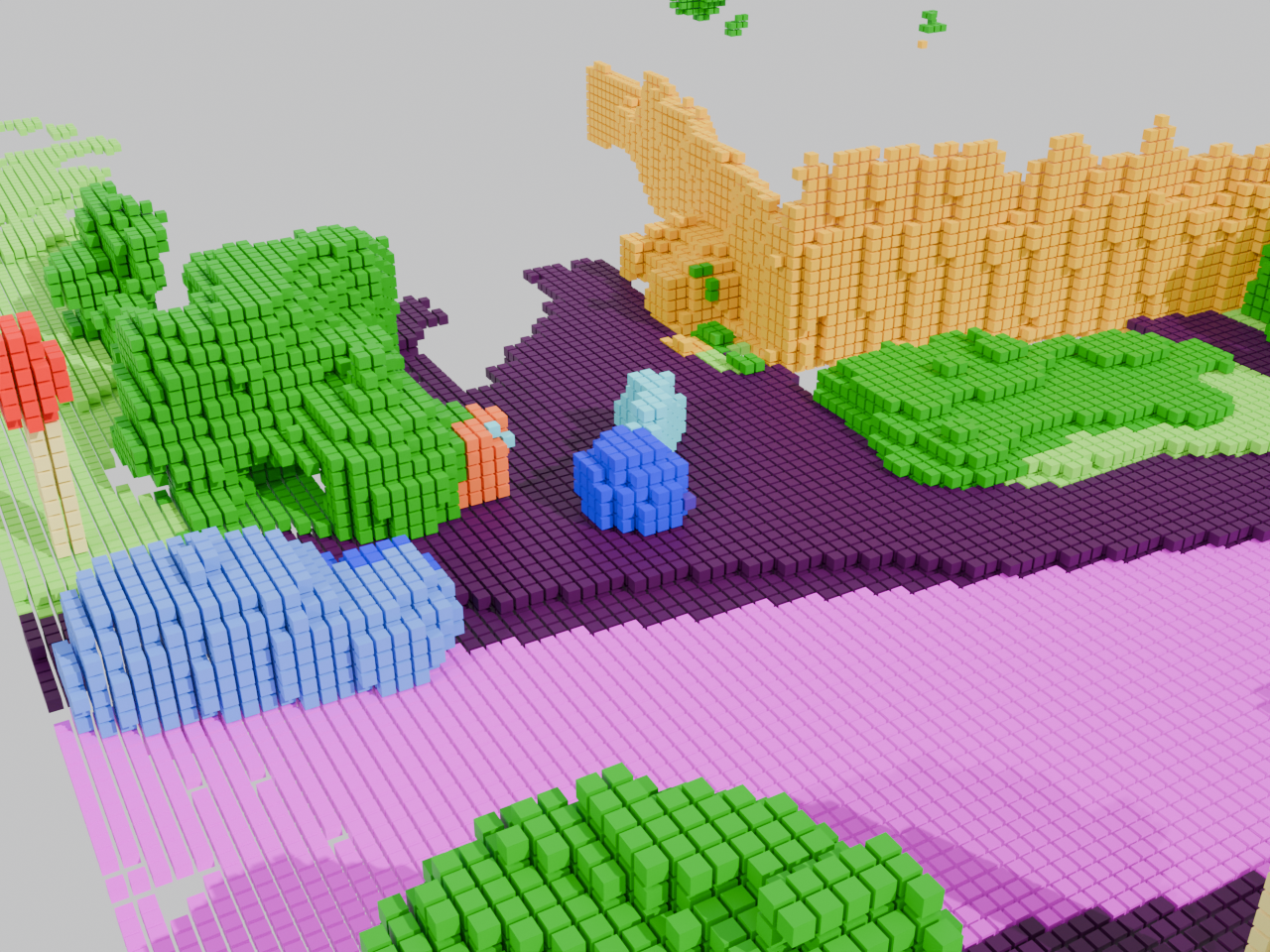} &
        \includegraphics[width=0.245\textwidth]{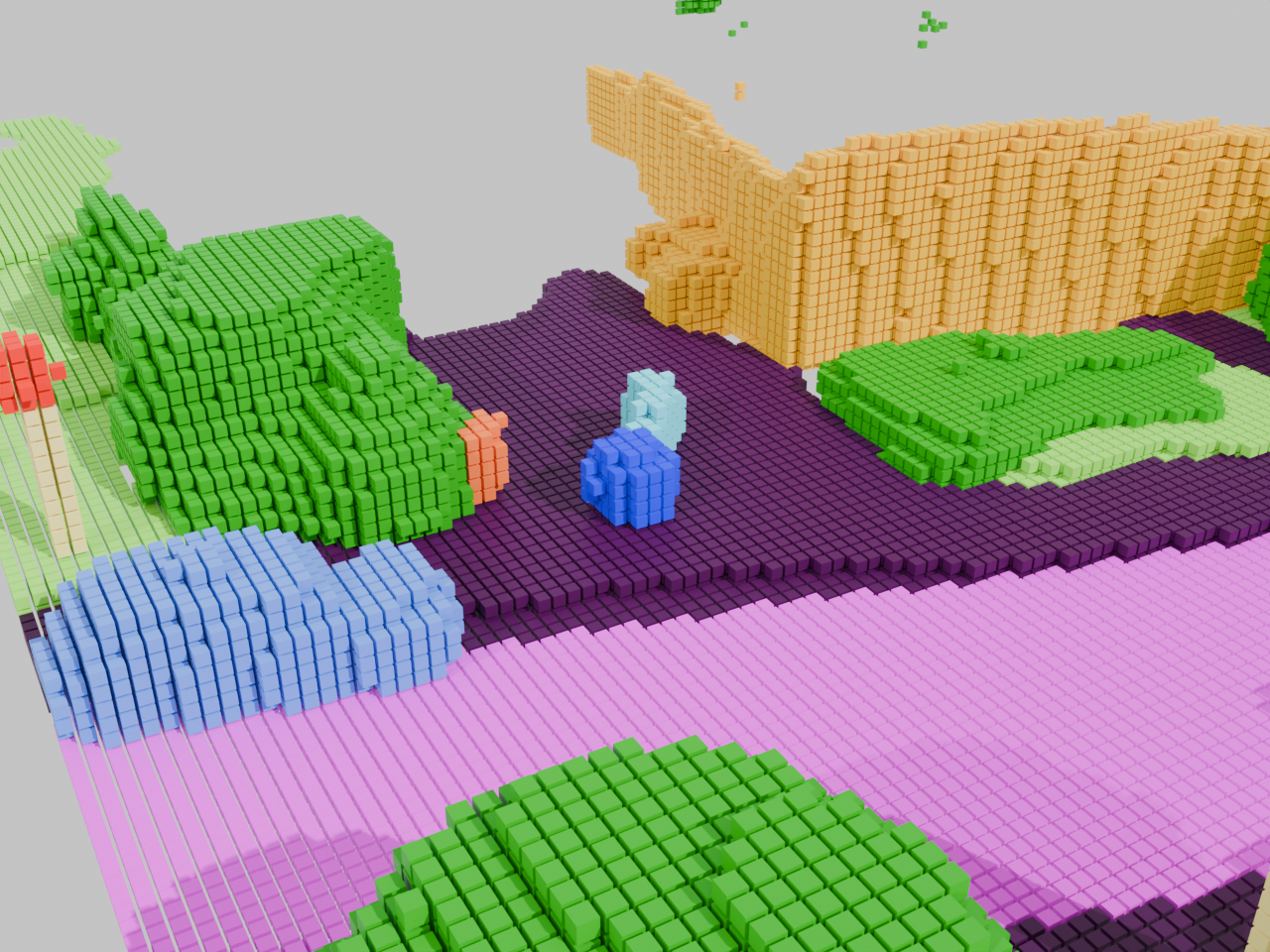} &
        \includegraphics[width=0.245\textwidth]{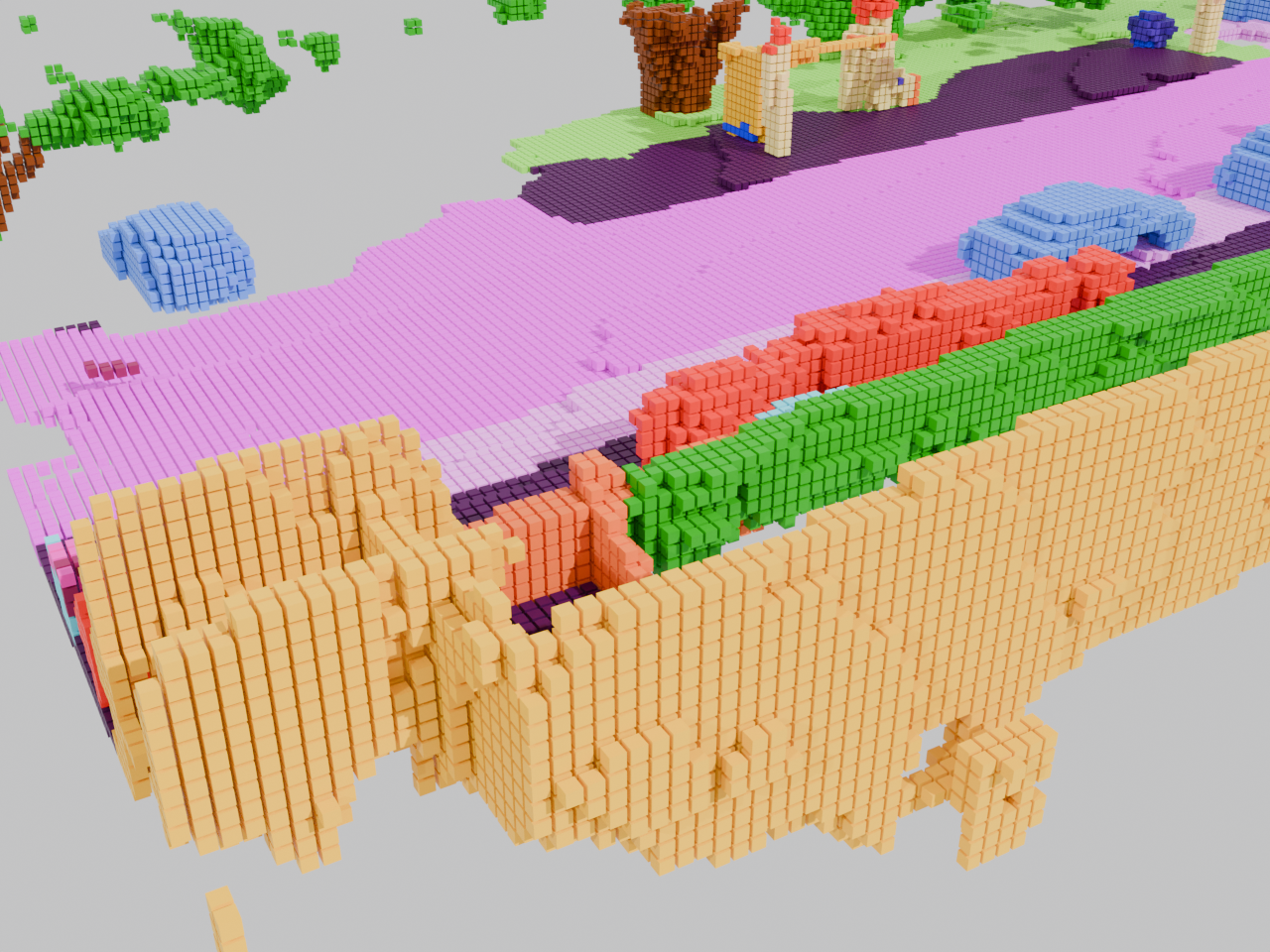} &
        \includegraphics[width=0.245\textwidth]{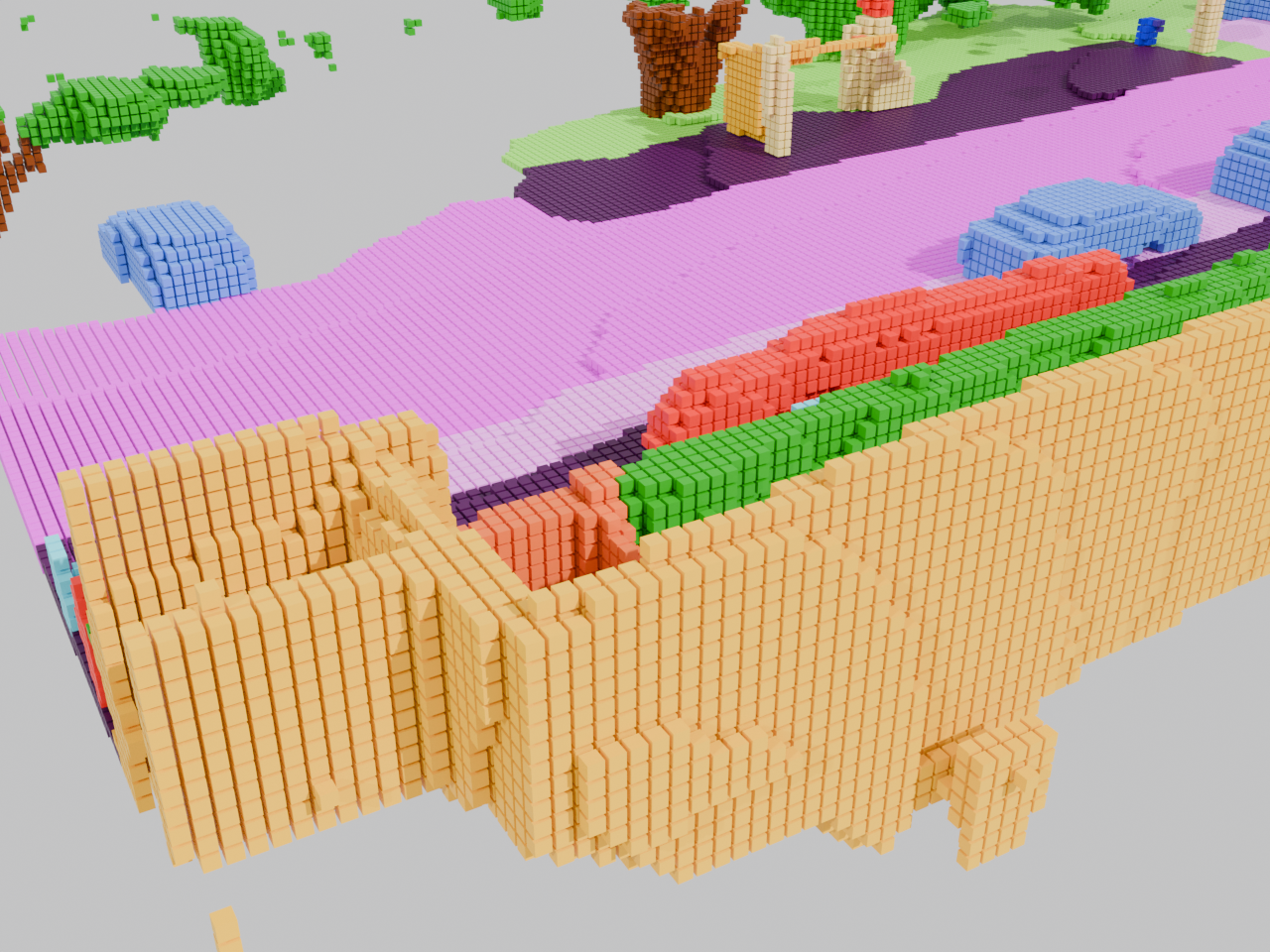} \\
    \end{tabular}
    \caption{\textbf{Six rare-class val scenes}, one per class, in descending order of $D_4$-TTA-vs-base IoU gain ($+40.1$ to $+20.1$). \textbf{Left}: SCPNet base; \textbf{right}: S\textsuperscript{2}D\textsuperscript{2} ($32$K multi-frame $+$ $D_4$ TTA), sharing one camera so differences are the predictions, not the framing. These are the scenes where the correction helps most; the regime where it \emph{hurts}, a weaker base whose source under-supplies rare classes, is \cref{fig:failure_vru}.}
    \label{fig:qualitative_gallery}
\end{figure*}

\begin{figure*}[!tbp]
    \centering
    \begin{tabular}{@{}c@{\hspace{0.5mm}}c@{}}
        \includegraphics[width=0.47\textwidth]{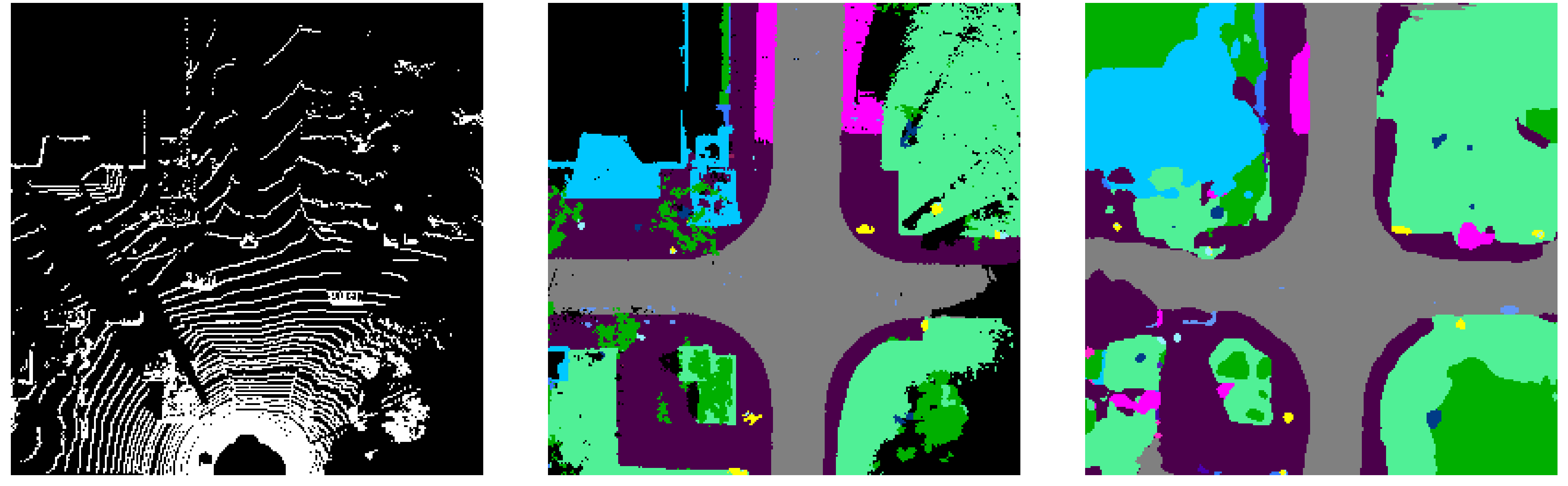} &
        \includegraphics[width=0.47\textwidth]{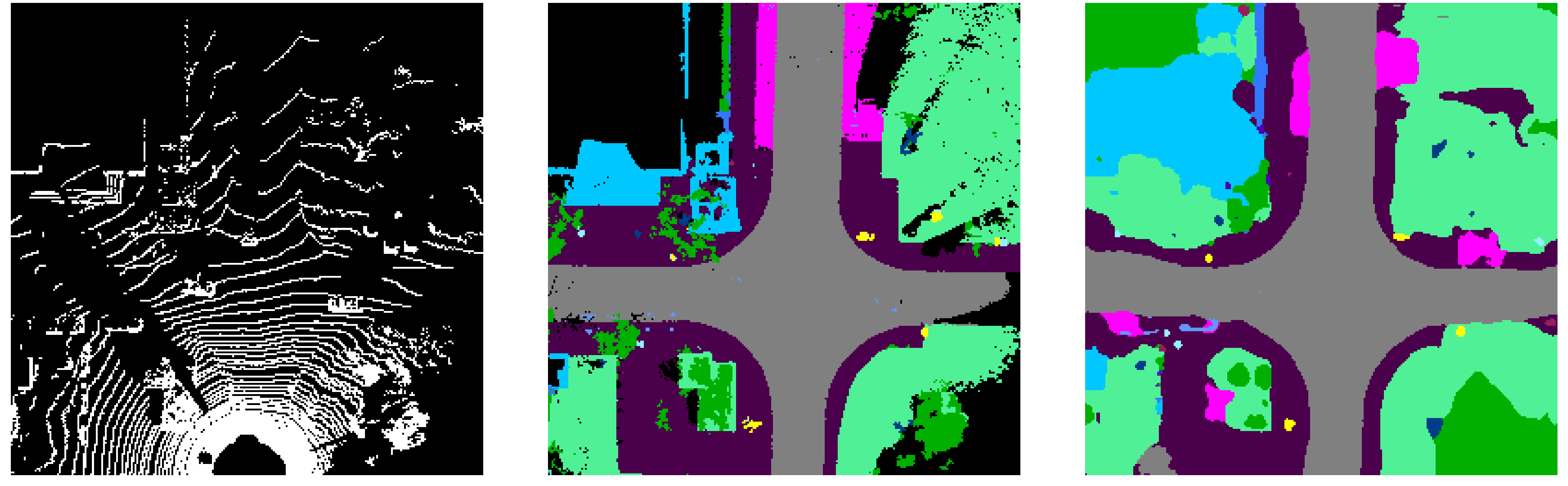} \\
        \includegraphics[width=0.47\textwidth]{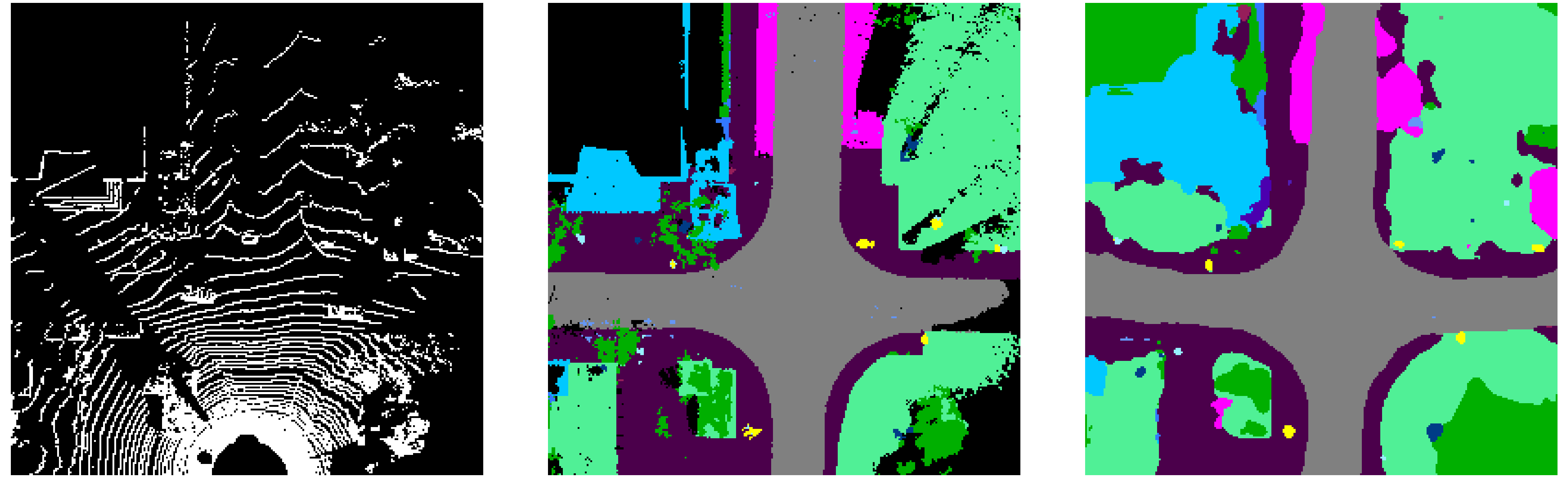} &
        \includegraphics[width=0.47\textwidth]{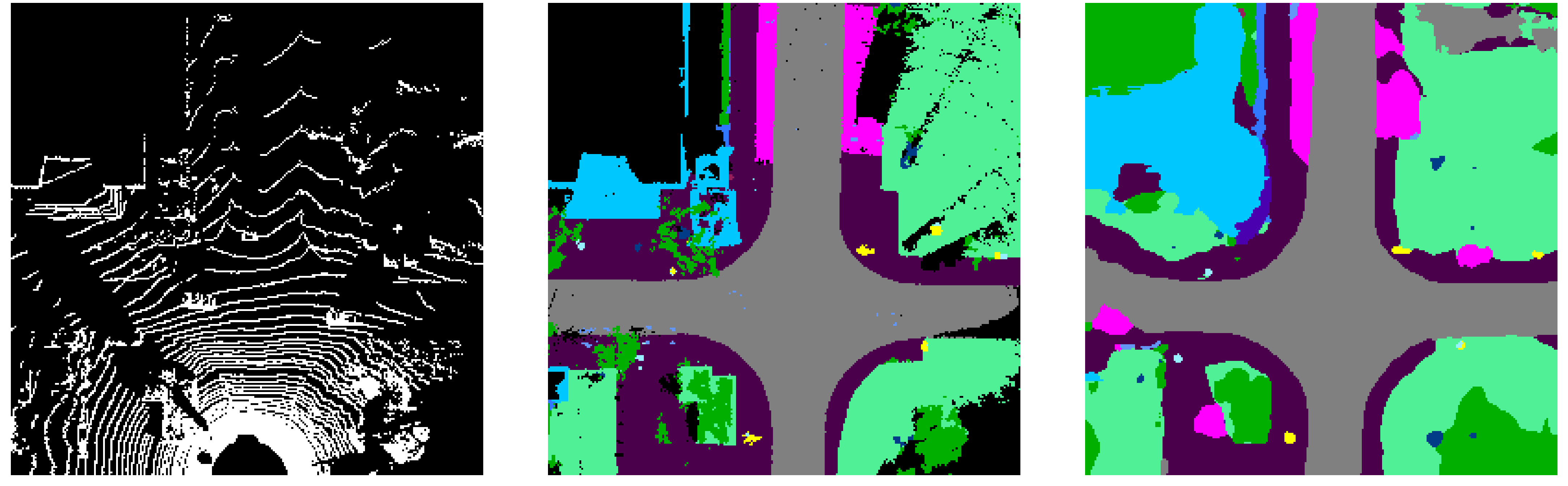} \\
    \end{tabular}
    \caption{\textbf{Qualitative BEV semantic predictions} on SemanticKITTI validation. Each panel is a different val seq-08 frame, not another view of one scene; within a panel, left to right: the sparse LiDAR input, the ground-truth BEV, and our predicted BEV. Colour key (this BEV renderer's own palette, not the voxel key): \textbf{grey}: road; \textbf{dark purple}: sidewalk; \textbf{magenta}: parking; \textbf{cyan}: building; \textbf{dark green}: vegetation; \textbf{light green}: terrain.}
    \label{fig:bev_qualitative}
\end{figure*}

The same operator also refines 2D BEV perception (\cref{fig:bev_qualitative}), though most of that headline is the projection rather than the operator. The pipeline reaches $\mathbf{36.1}\%$ BEV mIoU, of which $34.8$ is a parameter-free projection of the frozen 3D base (the topmost non-empty class per column, already $+7.8$\,pp over prior art), and S\textsuperscript{2}D\textsuperscript{2} adds $+1.3$. Swapping the 3D denoiser for a 2D variant and the source for the base-derived BEV, the same residual-transport machinery applies unchanged, and the result is $+9.1$\,pp over 2D S3CNet's $27.0\%$ (full comparison in Appendix~F).

\subsection{Where the Method Loses}
\label{sec:limits}
\label{sec:safety_metrics}

We close on the opposite question: where a reader should expect this operator to fail. Four limits are measured rather than argued.

\begin{figure*}[!tbp]
    \centering
    \setlength{\tabcolsep}{1.5pt}
    \renewcommand{\arraystretch}{0.6}
    \begin{tabular}{@{}ccc@{}}
        \includegraphics[width=0.318\textwidth]{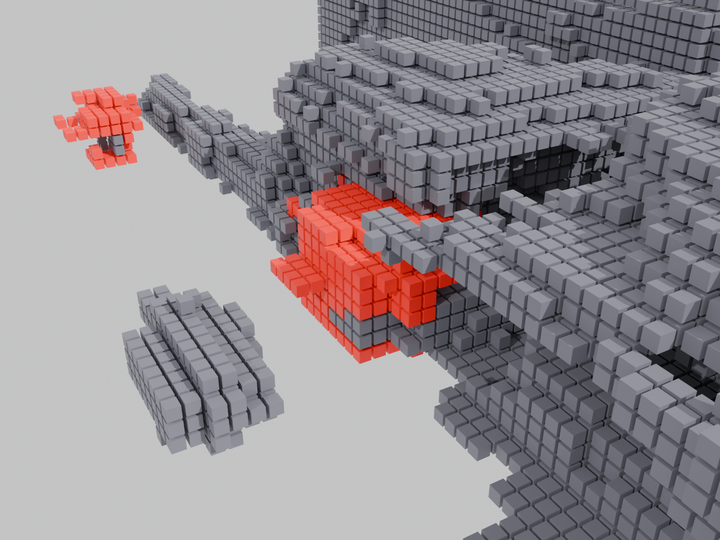} &
        \includegraphics[width=0.318\textwidth]{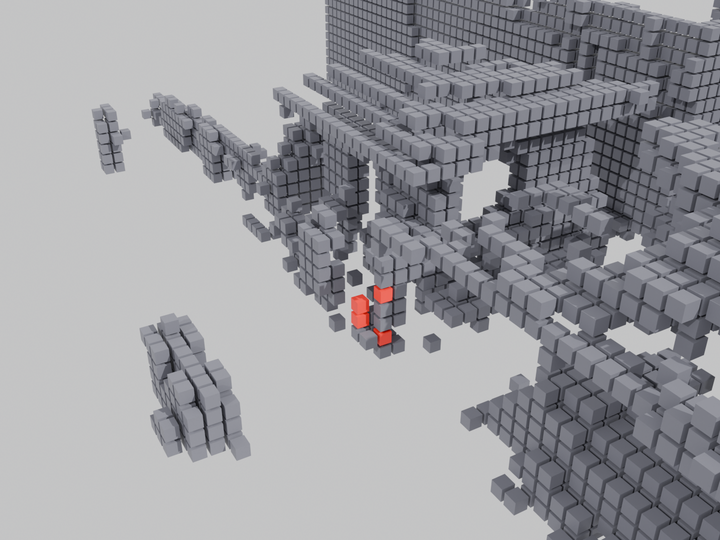} &
        \includegraphics[width=0.318\textwidth]{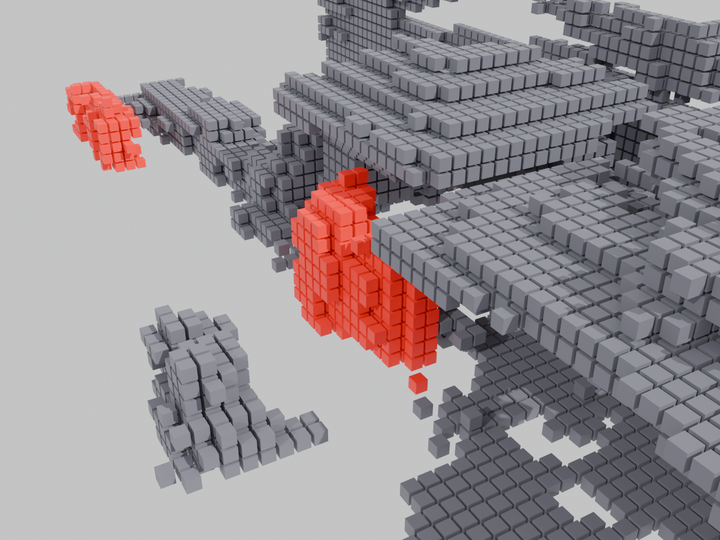} \\[2pt]
        \scriptsize (a) JS3C-Net base &
        \scriptsize (b) JS3C-Net $+$ S\textsuperscript{2}D\textsuperscript{2} &
        \scriptsize (c) ground truth \\
    \end{tabular}
    \caption{\textbf{Failure case: S\textsuperscript{2}D\textsuperscript{2} erases pedestrians on the JS3C-Net base} (val seq.~08, frame $001810$). Person and bicyclist voxels keep their semantic colour; the rest of the scene is muted to grey. Both pedestrians present in the ground truth (c) are recovered by the frozen base (a) and all but eliminated in the refined output (b), whose person IoU falls to $0.7\%$.}
    \label{fig:failure_vru}
\end{figure*}

\looseness=-1 \emph{It cannot exceed its guidance.} The generator inherits the ceiling of its conditioning, and the refiner inherits the ceiling of its source. Both were quantified above: predicted rather than oracle BEV collapses the two-wheeler classes while bulk geometry holds (\cref{fig:sgsc_conditioning}), and one correction step on the JS3C-Net base costs two of three VRU classes. Within the evaluated volume the failure mode is not hallucination but erasure. The never-observed volume the official metric excludes speaks to neither. On val seq.~$08$ frame~$001810$ one step \emph{erases} both pedestrians that the frozen JS3C-Net base had recovered, taking person IoU from $36.6$ to $0.7\%$, a source-bounded erasure typical of that base. Restricted to frames where the base recovers the class at all, one step lowers person IoU on $74.7\%$ of them and bicyclist IoU on $92.5\%$. On the SCPNet base the same two rates are $24.0\%$ and $23.5\%$, which is why we scope the transfer claim to voxel-grid-native sources (\cref{fig:failure_vru}; both per-frame breakdowns in Appendix~F).

\emph{The rarest class does not reproduce.} Motorcyclist's $+8.3$\,pp on val falls to $+0.3$ on retrain and to $+1.3$ against the published test row. It is the single class the safety story would most like to claim, and we do not claim it.

\emph{We do not lead on safety.} Among single-sweep rows S3CNet leads every VRU class, at VRU-IoU $32.6$ against our $21.6$, on an uneven profile whose completion IoU trails every other published test row. On the metric we introduce precisely because mIoU hides the tail, a $2021$ method is ahead of us, and closing that gap is open work.

\emph{Latency is not yet repaid.} The correction pass costs $107$\,ms, dropping the deployed base$+$refiner pipeline to $\mathbf{3.23}$\,FPS end-to-end on an idle H100, against the frozen base's $4.95$. Neither matches the sensor's $10$\,Hz sweep cadence~\cite{geiger2013kitti} (the refiner's marginal $9.33$\,FPS is an incremental pass, not a deployable rate). Charged for that latency at $T_w{=}1.0$\,s, DW-VRU-IoU moves $50.7\!\to\!49.9$ on val and $54.9\!\to\!49.7$ on test: one-step refinement trails the frozen base inside a one-second window. That is a statement about the equal-weight three-class mean, whose crossing sits past $T_w\approx1.2$\,s on val and $\approx3.2$\,s on test~\cite{aashto2018}; taken singly, no class crosses at all (window sweep in Appendix~C, extended limitations in Appendix~H). On test, therefore, the accuracy it buys does not yet pay for the time it costs. The $100$-step sampler is inadmissible under any budget, so the one-step reduction is what admits refinement into the latency envelope at all. A faster denoiser, not a better one, is what this result asks for next.

\section{Conclusion}
\label{sec:conclusion}

We recast outdoor LiDAR semantic scene completion from a discriminative mapping into a generative process (GSSC) driven by a single discrete-diffusion formulation across data synthesis (PS\textsuperscript{3}), base-free completion (SGSC), and base-conditioned refinement (S\textsuperscript{2}D\textsuperscript{2}). Conditioned on a frozen base, S\textsuperscript{2}D\textsuperscript{2} reaches $\mathbf{38.8\%}$ hidden-test mIoU at one step with no test-time augmentation, to our knowledge the best causal, single-sweep, single-sample leaderboard result to date. We also asked whether the operator is tied to the base it trained beside: it lifts three frozen bases spanning the field, retraining none and touching no architecture. The frozen checkpoint transfers zero-shot to SemanticPOSS and, on completion IoU, to SSCBench-KITTI360 (\cref{sec:crossdataset}), and the same generative view yields PS\textsuperscript{3}, a long-tail-rebalancing expansion of the SemanticKITTI training set. Generative modelling need not replace the discriminative pipeline to improve it: one diffusion formulation can supply its training data, complete a scene from noise, and correct what a deployed stack already produces.

\emph{Limitations.} The refiner cannot exceed its source: on a weaker base it can \emph{erase} a rare class rather than recover it, costing two of the three vulnerable-road-user classes on the JS3C-Net base (\cref{sec:limits}). Besides, its correction pass does not pay for its latency inside a one-second window. Our evidence for generality is two dense categorical tasks, 3D completion and BEV segmentation. Whether the correction reaches any output that is a per-element categorical simplex is untested rather than shown (Appendix~H).

\section*{Acknowledgments}
This work was supported by the National Natural Science Foundation of China (Grant 624B1006) and the Shanghai Science and Technology Committee (Grant 24511103900).


\bibliographystyle{IEEEtran}
\bibliography{main}

\ifdefined\GSSCarxiv\else
\begin{IEEEbiography}[{\includegraphics[width=1in,height=1.25in,clip,keepaspectratio]{material/photos/shi_chen_4x5.jpg}}]{Shi Chen}
received the BS degree in computer science from Fudan University, Shanghai, China, in 2026, where he worked on 3D computer vision, generative modeling, and autonomous driving. He is currently a graduate student with the Robotics Institute, Carnegie Mellon University, Pittsburgh, PA, USA. His current research interests include robot learning and mobile manipulation.
\end{IEEEbiography}

\begin{IEEEbiography}[{\includegraphics[width=1in,height=1.25in,clip,keepaspectratio]{material/photos/weifeng_ge_4x5.jpg}}]{Weifeng Ge}
received the PhD degree from the University of Hong Kong, in 2019. He is currently an associate professor with the College of Computer Science and Artificial Intelligence, Fudan University, Shanghai, China. His current research interests include computer vision, deep learning, artificial general intelligence, and humanoid robots.
\end{IEEEbiography}
\fi

\end{document}